\documentclass[11pt]{article}

\usepackage[final]{acl}

\usepackage{times}
\usepackage{latexsym}

\usepackage[T1]{fontenc}

\usepackage[utf8]{inputenc}

\usepackage{microtype}

\usepackage{inconsolata}

\usepackage{graphicx}

\usepackage{amsfonts} 
\usepackage{amssymb}  

\newcommand{\Ind}{\mathbb{I}}
\title{Beyond Black-Box Labels: Interpretable Criteria for Diagnosing Subjective NLP Tasks}

\author{
 \textbf{Nisrine Rair\textsuperscript{1,2}},
 \textbf{Alban Goupil\textsuperscript{1}},
 \textbf{Valeriu Vrabie\textsuperscript{1}},
 \textbf{Emmanuel Chochoy\textsuperscript{2}}
\\
\\
 \textsuperscript{1}CReSTIC, Université de Reims Champagne-Ardenne, Reims, France,\\
 \textsuperscript{2}Chochoy Conseil, Reims, France
\\
\href{mailto:nisrine.rair@univ-reims.fr}{\texttt{nisrine.rair@univ-reims.fr}}
\\
}

\begin{document}
\maketitle
\begin{abstract}

Subjective NLP datasets typically aggregate annotator judgments into a single gold label, making it difficult to diagnose whether disagreement reflects unclear criteria, collapsed distinctions, or legitimate plurality. We propose a \emph{schema-level diagnostic} for auditing expert-designed annotation 
schemas \emph{prior to} gold-label commitment, using only multi-annotator criterion judgments. The diagnostic separates two failure modes: unstable criteria with hard-to-operationalize boundaries, and systematic overlap 
that blurs the boundaries between mutually exclusive categories.
Applied to persuasive value extraction in commercial documents, we find that 
disagreement is not diffuse: instability concentrates in a few criteria, while nearly half of covered sentences activate multiple categories. These signals align with where domain experts disagree, yielding an evidence-based 
audit for tightening guidelines, revising category structure, or reconsidering the annotation paradigm.
Code and annotation data are publicly released 
\footnote{\url{https://github.com/NisrineRair/annotation-schema-diagnostic}}.
\end{abstract}

\section{Introduction}
    Natural Language Processing is undergoing a critical shift: many benchmarks increasingly measure subjective human judgments rather than factual accuracy \citep{uma_learning_2021, basile_we_2021}. Yet standard practice aggregates multiple judgments into a single gold label, collapsing disagreement into noise and misrepresenting subjective assessments as objective truth \citep{pavlick_inherent_2019}. In high-stakes settings, this opacity undermines reliability and can conceal biases, motivating more rigorous methods for designing and validating annotation schemas \citep{rottger_two_2022}.
    
    This challenge is especially acute in cold-start tasks, where no benchmarks or community norms exist. We study Persuasive Value Extraction (PVE) from commercial documents, a real application requiring categorization of sentences by procurement-relevant rationale. Following task decomposition practices \citep{sap_social_2020}, we operationalize this construct using expert-defined criteria. Yet multi-annotator labeling exhibits systematic disagreement, indicating that expert decomposition alone does not guarantee criteria are operationalizable in practice.
    The community has developed complementary tools: annotation paradigms to clarify labeling goals \citep{rottger_two_2022}, methods to quantify disagreement \citep{artstein_survey_2008, swayamdipta_dataset_2020}, and decomposition strategies to operationalize constructs \citep{sap_social_2020, ruggeri_let_2024}. Yet these advances often operate in isolation: Paradigms are often implicit, disagreement analyses treat schemas as fixed, and decompositions are rarely stress-tested \citep{uma_learning_2021, basile_we_2021}.
    
    This matters because once gold labels are constructed, schema failures are easily conflated with annotator noise and become expensive to diagnose or correct. When disagreement appears, its sources remain opaque. Specifically, disagreement may reflect (i) underspecified criteria, (ii) non-separable distinctions where criteria overlap, or (iii) legitimate plurality. Without criterion-level procedures to distinguish these sources, schema revision remains ad hoc. 
    To address this gap, we introduce a schema-level diagnostic that evaluates expert-designed schemas \emph{before} committing to gold labels, using only multi-annotator criterion judgments. The diagnostic isolates two actionable failure modes: (i) criterion instability, where criteria yield persistent borderline vote splits when engaged, and (ii) criterion overlap, where criteria co-activate and blur intended boundaries. Applied to PVE, disagreement is structured: instability concentrates in a few criteria, and overlap localizes to specific boundaries. These signals align with where domain experts later struggle, yielding an evidence-based audit for tightening guidelines, restructuring categories, or adopting an annotation paradigm that better reflects the inherent multi-dimensionality of the content.

\section{Related Work}

\paragraph{Subjectivity and Benchmark Reliability.}
    Evaluation benchmarks are central to NLP progress, yet their reliability is increasingly constrained by data quality and label consistency rather than model capabilities. Systematic label errors can distort evaluation \citep{northcutt_pervasive_2021, swayamdipta_dataset_2020}, a problem amplified by the field's shift toward inherently subjective tasks such as toxicity detection, stance detection, and subjectivity analysis. Common practice collapses subjectivity by aggregating judgments into a single majority-vote label, producing datasets built from subjective assessments yet presented as objective truth. This motivates explicit annotation paradigms that clarify what counts as ground truth.

\paragraph{Annotation Paradigms.}
    Annotation paradigms make normative and methodological commitments explicit. In a prescriptive paradigm, the goal is to reduce variation by enforcing a single interpretation. In a descriptive paradigm, the goal is to measure variation across legitimate judgments \citep{aroyo_truth_2015}. A perspectivist view models structured viewpoints, treating disagreement as a reflection of coherent, underlying perspectives \citep{basile_toward_2023}. The perspectivist shift has inspired methods learning from disagreement, including learning from soft-label distributions \citep{uma_learning_2021}, analyzing training dynamics \citep{swayamdipta_dataset_2020}, and modeling annotator-specific parameters \citep{mostafazadeh_davani_dealing_2022}. However, these approaches typically assume the schema is adequate and do not test whether disagreement reflects underspecified criteria, structural overlap, or legitimate plurality. Our diagnostic instead tests schema adequacy, helping practitioners decide whether to tighten guidelines, redesign distinctions, or adopt multi-perspective annotation.

\paragraph{Measuring and Analyzing Label Variation.}
Operationalizing any annotation paradigm requires measuring and characterizing label variation. Existing methods fall into three broad families. \textbf{Global agreement metrics}, such as Cohen's $\kappa$ and Krippendorff's $\alpha$, quantify overall reliability but do not localize whether low agreement reflects annotator errors, inherent subjectivity, or underspecified guidelines \citep{artstein_survey_2008, ruggeri_let_2024, plank_problem_2022}. \textbf{Instance-level localization} treats disagreement as a property of particular examples, using methods such as CrowdTruth \citep{aroyo_truth_2015} and Dataset Cartography \citep{swayamdipta_dataset_2020} to localize contentious items. \textbf{Annotator attribution} models annotator behavior, either as noisy channels characterized by a confusion matrix \citep{paun_comparing_2018, dawid_maximum_1979} or by adapting models to capture individual perspectives \citep{ignatev_hypernetworks_2025, plepi_unifying_2022}. Conceptual frameworks further distinguish sources of disagreement: instance ambiguity, annotator subjectivity, and task underspecification \citep{basile_toward_2023}. Because these methods operate on labels after the schema is set, they primarily organize outputs around items and annotators, rather than testing whether criteria are applied consistently or whether intended distinctions are empirically separable.

\paragraph{Task Operationalization and Schema Decomposition.}
To address subjectivity, researchers increasingly focus on task operationalization, decomposing high-level constructs into concrete labeling frameworks. Documentation practices such as Data Statements \citep{bender_data_2018} and Datasheets for Datasets \citep{gebru_datasheets_2021} surface design choices, but they often leave unclear whether different annotators can apply the resulting distinctions consistently. Recent work has also made guidelines and disagreement patterns central objects of analysis. The Guideline-Centered Annotation Methodology \citep{ruggeri_let_2024} evaluates outcomes against a fixed expert-defined schema, emphasizing guideline adherence. Disagreement-centric frameworks such as CrowdTruth \citep{aroyo_truth_2015} and analyses such as Dataset Cartography \citep{swayamdipta_dataset_2020} treat disagreement as signal about item difficulty or annotator variation under an assumed label inventory. Domain-specific approaches further anchor criteria in external normative frameworks \citep{jikeli_antisemitic_2023} or apply semantic componential analysis \citep{korre_untangling_2025, salminen_anatomy_2018}. Overall, these approaches justify schema adequacy through expert design, but provide limited empirical stress-testing of whether criteria remain operationally consistent or boundaries remain separable in practice. 
Across these lines of work, the annotation schema is typically treated as a fixed input, while disagreement is primarily characterized at the level of items or annotators. Consequently, existing methods rarely distinguish whether disagreement originates in (i) underspecified criterion boundaries, which is often addressable through tighter guidance, or (ii) structural overlap, where intended distinctions systematically co-activate and require category redesign or a different annotation paradigm. Our work addresses this gap with schema-level evaluation from \textbf{criterion-level annotations}, directly auditing criterion stability and cross-criterion separability during schema development.

\section{Methodology: Task Diagnosis}
\label{sec:method}
We propose a task-agnostic, schema-level diagnostic for subjective classification that does not assume a single correct label. Instead of inferring a latent ``true'' label, we analyze multi-annotator judgments as repeated applications of written criteria. Our diagnostic evaluates whether a schema behaves stably and distinctly in practice, thereby probing its operationalization, particularly the clarity of its boundaries and the separability of its categories.
Concretely, we test whether the schema yields distinctions that are: (i) \textbf{applied consistently} across repeated invocations, and (ii) \textbf{empirically separable} in joint judgments. The definition of separability depends on the intended annotation paradigm: mutually exclusive taxonomies require clear separability, whereas overlap may be acceptable or even expected in multi-label or multi-perspective settings.
A criterion exhibits instability when, conditional on annotator engagement, 
judgments frequently result in intermediate vote splits rather than converging toward unanimity. Depending on the annotation paradigm, this pattern may indicate an underspecified boundary or reflect legitimate annotator variation, and the diagnostic surfaces this signal without prescribing its interpretation. 
In contrast, a schema exhibits non-separability when multiple criteria (and their induced categories) overlap on the same unit at non-trivial rates. In a single-label setting, this overlap makes it impossible to select a single, unambiguous label without an explicit tie-breaking policy, though in a descriptive or multi-label paradigm, such co-activation may instead reflect the genuine 
multi-dimensionality of the content, with criterion-level overlap further localizing which specific boundaries are under pressure beyond what coarse category disagreement alone reveals.
\paragraph{Task schema.}
We formalize a task schema \footnote{See Table~\ref{tab:notation} in Appendix~\ref{app:notation} for a notation reference.} as $\mathcal{T}=(\mathcal{C},\mathcal{Q},\mu)$, where $\mathcal{C}=\{c_0,c_1,\dots,c_{C}\}$ is the set of categories. We designate $c_0$ as the \textbf{non-target category} (e.g., ``no persuasive value'' in our setting, analogous to ``neutral'' or ``non-offensive'' in other tasks), and treat the remaining $C$ categories $\{c_k\}_{k=1}^{C}$ as substantive outcomes
\footnote{Depending on the task, the non-target outcome may be represented explicitly as a dedicated category or left implicit as the absence of any substantive label.}.
The set $\mathcal{Q}=\{q_1,\dots,q_Q\}$ contains $Q$ expert-defined binary criteria, each phrased as a yes/no question answered for each unit to test for a specific semantic signal. The mapping $\mu:\mathcal{Q}\rightarrow \mathcal{C}$, the criterion-to-category 
assignment, associates each criterion with the category it is intended to support. 
For each substantive category $c_k$, we denote its supporting criteria by 
$\mathcal{Q}_k=\{q\in\mathcal{Q}\mid \mu(q)=c_k\}$, the criteria whose 
positive judgment is taken as evidence for $c_k$, treating both $\mathcal{Q}$, the set of criteria assigned to categories, and $\mu$, 
their category assignments, as revisable design choices to be evaluated through our diagnostic audits during schema development.

\paragraph{Annotation setup.}
Given a corpus $\mathcal{S}=\{s_1,\dots,s_S\}$ of annotation units (e.g., sentences) and a panel of annotators $\mathcal{A}=\{a_1,\dots,a_A\}$ (human, LLM, or hybrid), each annotator provides a yes/no response for each criterion $q$ on each unit $s$. This yields a binary response tensor $\mathbf{Y}\in\{0,1\}^{S\times A\times Q}$, where $y_{saq}=1$ if annotator $a$ marks criterion $q$ as present in unit $s$, and $0$ otherwise. For each unit--criterion pair $(s,q)$, we compute the positive vote count:
\begin{equation}
\label{eq:votes}
v_{sq}=\sum_{a=1}^{A} y_{saq},
\end{equation}
which implies $v_{sq}\in\{0,\dots,A\}$, where $v_{sq}=0$ means no annotator 
marked criterion $q$ as present for unit $s$, and $v_{sq}=A$ means full 
unanimity.
Because criteria differ widely in how often they are triggered, a sparsely 
activated criterion will appear highly stable simply because annotators 
consistently agree on its absence, masking genuine boundary ambiguity on 
the units where it actually applies, unless universal absence is itself 
the expected and theoretically motivated outcome. To avoid masking boundary behavior, we restrict analysis to units where at least $t$ annotators marked the criterion as present, focusing on cases where the criterion is meaningfully engaged. Formally, for a criterion $q$ and threshold $t\in\{0,1,\dots,A\}$, we define the \textbf{focus set} as:
\begin{equation}
\label{eq:omega}
\Omega_{q,t}=\{\, s\in \mathcal{S} \mid v_{sq}\ge t \,\}.
\end{equation}
Intuitively, $\Omega_{q,t}$ is the \emph{relevant scope} for evaluating 
criterion $q$: the units for which $q$ is meaningfully engaged.
When $t\ge 1$, $\Omega_{q,t}$ restricts analysis to \emph{engaged} cases. 
The choice of $t$ controls the selectivity of the analysis: setting $t=1$ 
includes all units where at least one annotator marked criterion $q$ as 
present, while $t=0$ recovers the full corpus $\mathcal{S}$. Higher values 
of $t$ impose stricter engagement, focusing on units where multiple 
annotators agree the criterion applies.

\paragraph{Criterion stability.}
Stability refers to whether a criterion $q$ yields consistent presence 
judgments when it is engaged. The size of the focus set $\Omega_{q,t}$ 
is itself informative: a large focus set indicates a frequently triggered 
criterion, while a small one signals a rare or narrow signal. We formalize 
this as the \textbf{activation rate}:
\begin{equation}
\label{eq:act}
\mathrm{Act}_t(q)=\frac{|\Omega_{q,t}|}{|\mathcal{S}|},
\end{equation}
which gives the proportion of corpus units where criterion $q$ is engaged, capturing both its definitional scope and how prevalent the corresponding signal is in the corpus. A low activation rate may indicate a rare but task-relevant signal, while a high 
rate suggests a pervasive one, both informative about how the schema behaves in practice. Second, given the focus set $\Omega_{q,t}$, the \textbf{conditional vote 
distribution} $\pi_q(\cdot\mid t)$ provides, for each criterion, a 
distribution over agreement types: from full unanimity, where all annotators 
agree on presence, to near-equal splits, where the criterion sits at a 
boundary. Formally, for each level of agreement $k \in \{t, \dots, A\}$:
\begin{equation}
\label{eq:pi}
\pi_q(k\mid t)=\frac{|\{\, s\in\Omega_{q,t} \mid v_{sq}=k \,\}|}{|\Omega_{q,t}|},
\end{equation}
where the numerator counts engaged units receiving exactly $k$ positive 
votes and the denominator is the total number of engaged units, with $k$ 
ranging from minimal engagement ($k=t$) to full unanimity ($k=A$). 
We summarize $\pi_q(\cdot\mid t)$ with ambiguity-based metrics in 
Section~\ref{sec:engrule}. Concentration near $k=A$ indicates consistent 
application, while mass at intermediate values signals boundary ambiguity. 
This captures the extent of disagreement but does not, by itself, distinguish diffuse uncertainty from stable splits into annotator subgroups, a distinction we return to via overlap.

\paragraph{Criterion separability.}
Stability evaluates each criterion in isolation, but does not reveal whether 
criteria fire independently or tend to co-activate on the same units. 
Co-activation is structurally informative regardless of the annotation 
paradigm: it reveals where the schema's intended distinctions break down 
in practice, whether this signals a design flaw or the genuine 
multi-dimensionality of the content. We therefore measure the joint 
behavior of criteria across the corpus using the same focus sets 
$\Omega_{q,t}$.

For each unit $s$, we identify which criteria are simultaneously engaged 
by aggregating annotator votes: a criterion $q$ is considered engaged for 
unit $s$ if at least $t$ annotators marked it as present. $\Gamma_{s,t}$ 
collects all such criteria:
\begin{equation}
\label{eq:gamma}
\Gamma_{s,t}=\{\, q \mid s\in \Omega_{q,t} \,\},
\end{equation}
and $|\Gamma_{s,t}|$ counts how many criteria fire simultaneously for 
unit $s$. If criteria were perfectly separable, each unit would activate 
exactly one criterion. Units where $|\Gamma_{s,t}|>1$ therefore signal potential co-activation between criteria. To measure 
how systematically two criteria co-occur and distinguish symmetric 
entanglement from subset-like behavior, we define a directed conditional 
overlap:
\begin{equation}
\label{eq:condov}
\mathrm{CondOv}_t(q\rightarrow q')=\frac{|\Omega_{q,t}\cap \Omega_{q',t}|}{|\Omega_{q,t}|},
\end{equation}
which estimates the empirical conditional probability that $q'$ is engaged 
when $q$ is engaged. Asymmetry is informative: high 
$\mathrm{CondOv}_t(q\rightarrow q')$ with low 
$\mathrm{CondOv}_t(q'\rightarrow q)$ suggests that $q$ tends to occur 
as a subset of $q'$, indicating potential redundancy or confounding 
rather than mutual overlap. Detecting such 
asymmetries is important because it reveals latent semantic structure 
in the schema: criteria may form implicit hierarchies where one subsumes 
another, signaling that the schema's intended distinctions may not be 
operating at the same level of specificity.

\paragraph{Mapping to categories.}
Criterion co-activation captured by $\Gamma_{s,t}$ does not distinguish 
whether co-occurring criteria belong to the same category or to distinct 
ones. Yet this distinction matters: criteria within the same category may tend to co-occur by construction, while co-activation across distinct categories 
signals that the schema's intended boundaries may not hold in practice. 
To make this distinction explicit, we lift engagement to the category 
level using the mapping $\mu$. For each substantive category $c_k$, we define a binary indicator that captures whether the category is active for unit $s$, that is, whether at least one of its supporting criteria is engaged:
\begin{equation}
\label{eq:gik}
g_{sk,t}=\Ind\!\left(\exists\, q \in \mathcal{Q}_k:\ s \in \Omega_{q,t}\right),
\end{equation}
where $\Ind(\cdot)$ equals $1$ if its argument is true and $0$ otherwise. 
Intuitively, $g_{sk,t}=1$ means the schema fires category $c_k$ for 
unit $s$, and $g_{sk,t}=0$ means it does not.

Summing over all substantive categories, $m_{s,t}=\sum_{k=1}^{C} g_{sk,t}$ 
counts how many distinct categories are simultaneously active for unit $s$. 
We call a unit \emph{covered} at threshold $t$ if $m_{s,t}\ge 1$, meaning 
the schema fires at least one substantive category for that unit.
Because $c_0$ is defined implicitly as the absence of any substantive 
category, including uncovered units would dilute overlap rates with units 
where the schema simply does not fire. We therefore measure, among units where the schema fires, the fraction 
for which it fires on more than one category simultaneously, that is, 
the rate of cross-category boundary blurring conditional on coverage:
\begin{equation}
\label{eq:overlap-cat-cov}
\mathrm{Overlap}_{\mathrm{cat}\mid \mathrm{cov},\,t}=
\frac{\sum_{s\in\mathcal{S}} \Ind\!\left(m_{s,t}\ge 2\right)}
     {\sum_{s\in\mathcal{S}} \Ind\!\left(m_{s,t}\ge 1\right)},
\end{equation}
where the numerator counts covered units activating at least two categories, 
and the denominator counts all covered units. A high value indicates that 
the schema frequently fires on multiple categories simultaneously, making 
single-label assignment ambiguous without an explicit tie-breaking policy.
Finally, units with large $|\Gamma_{s,t}|$ indicate many criteria firing 
at once, and units with $m_{s,t}\ge 2$ indicate cross-category boundary 
blurring, providing instance-level diagnostics for targeted schema revision.

\section{Case Study: Diagnosing the PVE Schema}
\subsection{Persuasive Value Extraction}
\label{sec:pve}
    Having introduced a task-agnostic schema diagnostic, we instantiate it on \textit{Persuasive Value Extraction} (PVE), a real-world formalization effort in commercial document analysis \cite{chochoy2025msmkc}. We do not present PVE as a benchmark or claim the schema is definitive. 
    Instead, it serves as a cold-start setting with no established community norms, making it well suited to demonstrate how disagreement can stress-test a schema and guide revision.
    PVE targets statements that articulate a benefit, advantage, or requirement relevant to purchase justification, as opposed to purely descriptive content. Domain experts initially formulated PVE as a sentence-level \textbf{single-label} classification task to support procurement workflows that require mutually exclusive categories. The task comprises four categories:
    \begin{itemize}
      \setlength{\itemsep}{-5pt}
      \item \textbf{Non-Persuasive} ($c_0$): descriptive content with no explicit value claim.
      \item \textbf{Performance \& Efficiency} ($c_1$): cost, efficiency, or measurable outcomes.
      \item \textbf{User Experience \& Brand Value} ($c_2$): well-being, perceived quality, reputation, or appeal.
      \item \textbf{Obligation \& Safety} ($c_3$): compliance, requirements, risk reduction, or security.
    \end{itemize}

\begin{figure}[ht]
  \centering
  \includegraphics[width=0.95\columnwidth]{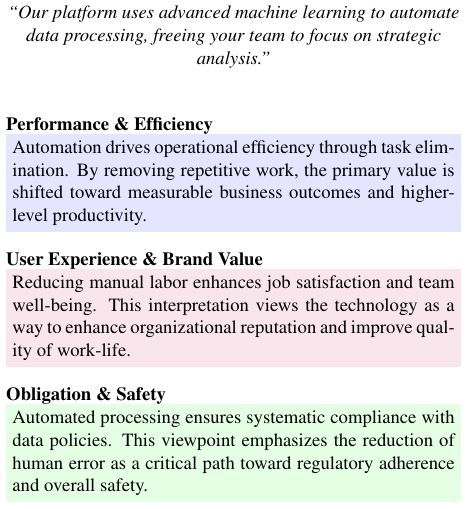}
      \caption{PVE boundary case: the same sentence supports multiple defensible readings tied to different organizational priorities, yielding conflicting single-label assignments.}
     \label{fig:pve_boundary_example}
\end{figure}

    Initial annotation revealed systematic disagreement. Persuasive value admits multiple defensible readings: the same sentence can be interpreted as emphasizing different advantages (financial, operational, compliance, or reputational), and annotators may weigh these advantages over different time horizons. In particular, what looks like a short-term advantage can be judged less persuasive when potential longer-term downsides are taken into account, producing conflicting but reasonable single-label assignments. Figure~\ref{fig:pve_boundary_example} shows a boundary case where experts split across categories. Additional task details and examples are provided in Appendix~\ref{app:task-examples}.

\paragraph{Schema instantiation.}
To instantiate the PVE schema for diagnosis, we define $\mathcal{T}=(\mathcal{C},\mathcal{Q},\mu)$ with $\mathcal{C}=\{c_0,c_1,c_2,c_3\}$. Domain experts decomposed the persuasive categories ($c_1$--$c_3$) into three yes/no criteria each, reflecting an initial consensus on a small set of core signals per category rather than a definitive decomposition. This yields $M=9$ criteria, $\mathcal{Q}=\{q_1,\dots,q_9\}$. The mapping $\mu:\mathcal{Q}\rightarrow \mathcal{C}$ assigns each criterion to exactly one category, with $\{q_1,q_2,q_3\}\mapsto c_1$, $\{q_4,q_5,q_6\}\mapsto c_2$, and $\{q_7,q_8,q_9\}\mapsto c_3$. The criteria provide a higher-resolution probe of the single-label PVE schema: they allow us to test whether the intended category distinctions are applied consistently and remain separable in annotator behavior, or whether multi-faceted value statements systematically trigger multiple signals. Full criterion wording, decision guidance, and the iterative calibration process through which these criteria were developed are detailed in Appendix~\ref{app:criteria}.

\subsection{Experimental Setup}
\label{sec:experimental-setup}

\paragraph{Corpus and data collection.}
Our dataset $\mathcal{S}$ comprises $|\mathcal{S}|=4{,}701$ sentences (annotation units $s$) extracted from 65 B2B commercial documents spanning five clients and multiple sectors (e.g., cybersecurity, mobility, regulatory affairs, infrastructure, environmental services).
Documents are primarily in French, with English technical terminology and a small number of fully English documents.
For schema-level diagnosis, coverage of diverse document types is more informative than scale.
Text was extracted from PDFs using layout-aware processing to recover reading order, followed by rule-based sentence segmentation.
We avoid heavy cleaning of extraction artefacts, as they are part of the real input distribution and influence how criteria are applied.
Additional corpus statistics are provided in Appendix~\ref{app:dataset-details}.

\paragraph{Annotation protocol.}
To instantiate $\mathbf{Y}$ at scale, we use a panel of LLMs as a reproducible 
probe of how written criteria behave under repeated application: each criterion 
is queried independently using a fixed template. We use this LLM panel as a 
\emph{stress-test instrument} for the written criteria, to expose where the 
schema is unstable or non-separable, rather than as a proxy for human ground 
truth. Since the goal is schema auditing rather than label collection, 
LLM-specific variation is treated as any annotator variation would be in a 
multi-annotator study: a measurable signal rather than noise, whose structure 
can be analyzed, decomposed, and interrogated independently of its source. 
The diagnostic is agnostic to annotator type (human, LLM, or hybrid): here, 
LLMs are used to surface criterion-level instability and co-engagement patterns, 
and we validate key findings against expert human annotations on a subset of 
sentences. Each model was independently queried on the corpus $\mathcal{S}$ 
using the same instruction prompt\footnote{Full prompt text is in 
Appendix~\ref{app:prompts}, response handling details are in 
Appendix~\ref{app:response-cleaning}, and model panel details are in 
Appendix~\ref{app:model-panel}.} and the $Q=9$ yes/no PVE criteria $q$, 
with deterministic decoding (temperature $=0$) and a short output budget 
(max tokens $=3$). We use a diverse panel of $A=5$ models as annotators 
drawn from distinct model families: \texttt{gpt-4.1-mini}, \texttt{gpt-4.1}, 
\texttt{llama-3.3-70b}, \texttt{mistral-large-2411}, and \texttt{Qwen-2.5-72B}. 
Outputs were parsed into binary criterion decisions. Two sentences were removed 
due to persistent formatting failures, yielding 
$\mathbf{Y}\in\{0,1\}^{4{,}699\times 5\times 9}$.

\paragraph{Human validation subset.}
To anchor diagnostic patterns in human judgment, five domain experts with 
procurement expertise annotated a validation subset of 500 units: 389 unique 
sentences stratified by industry sector, plus 111 repeats to assess reliability. 
Experts assign exactly one category from $\mathcal{C}=\{c_0,c_1,c_2,c_3\}$ 
per unit, matching the intended single-label task design. This subset is used 
only to test whether criterion-level instability and overlap correlate with 
category-level expert disagreement and recurrent boundary confusions. We asked 
experts to assign categories directly rather than answer all $Q=9$ criteria, 
as criterion-level annotation would be substantially more time-consuming. 
Moreover, since the criteria were defined by domain experts to operationalize 
the categories, category-level judgments better reflect their intended holistic 
reading of persuasive value in procurement settings. These annotations are 
therefore treated as an external validity check rather than ground truth: 
alignment between diagnostic signals and expert disagreement patterns confirms 
that instability and overlap reflect schema-level structure rather than 
LLM-specific artifacts. Details appear in Appendix~\ref{app:annotation-details}.

\paragraph{Focus-set thresholding.}
\label{sec:engrule}
PVE criteria are sparse because many units are purely descriptive and map to the non-target category $c_0$.
We therefore use $t=1$ in the main analysis and define the focus set
$\Omega_{q,1}=\{\, s\in\mathcal{S}\mid v_{sq}\ge 1 \,\}$,
i.e., units where criterion $q$ receives at least one positive vote.
We default to $t=1$ because higher thresholds condition on majority agreement and would systematically exclude borderline or minority-but-defensible applications, precisely the cases most informative about boundary clarity.
Robustness to stricter engagement regimes is reported in Appendix~\ref{app:robustness-rules}.
Conditioning on $\Omega_{q,1}$ ensures that $\mathrm{Act}_{1}(q)$ and $\pi_q(\cdot\mid 1)$ characterize criterion behavior \emph{when it is engaged}, rather than being dominated by universal absence.

We summarize the distribution $\pi_q(\cdot \mid 1)$ (Eq.~\ref{eq:pi}) with three interpretable rates for the $A=5$ case, partitioning the vote distribution into three diagnostic zones:
\begin{itemize}
  \setlength{\itemsep}{-5pt}
  \item \textbf{UY (unanimous yes)}: $\pi_q(5\mid 1)$.
  \item \textbf{AS (asymmetric split)}: $\pi_q(4\mid 1)+\pi_q(1\mid 1)$, capturing 4--1 and 1--4 vote splits.
  \item \textbf{NT (near-tie)}: $\pi_q(3\mid 1)+\pi_q(2\mid 1)$, capturing boundary pressure via 3--2 and 2--3 splits.
\end{itemize}

While the diagnostic procedure is task-agnostic, the instability and overlap 
patterns reported below are properties of this PVE instantiation and should 
not be interpreted as universal regularities.

\section{Results and Discussion}
\label{sec:results}

\begin{figure}[th]
  \centering
  \includegraphics[width=0.99\columnwidth]{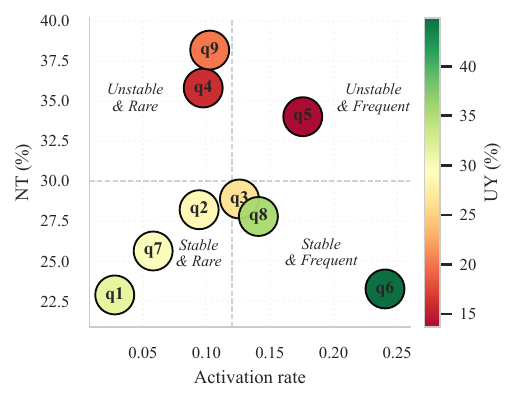}
\caption{Stability landscape at $t=1$. Each criterion is positioned by activation rate $\mathrm{Act}_1(q)$ (x-axis) and near-tie rate $\mathrm{NT}$ (y-axis), computed over the focus set $\Omega_{q,1}$. Color encodes unanimity $\mathrm{UY}$.}
  \label{fig:stability_landscape}
\end{figure}

\paragraph{Subjectivity is diagnosable: criteria exhibit systematic instability.}
A core goal of schema diagnosis is to test whether each criterion acts as a stable measurement instrument when invoked. With $t=1$, Table~\ref{tab:criterion-stability} and Fig.~\ref{fig:stability_landscape} show strong criterion-specific variation in activation and conditional vote structure. Some criteria are comparatively crisp: \textbf{$q_6$} (\emph{Perceived Quality}) is frequently engaged ($\mathrm{Act}_{1}(q_6)=24.0\%$) and often yields unanimity (UY $=44.9\%$), consistent with clearer operational boundaries. Others show persistent boundary ambiguity: \textbf{$q_9$} (\emph{Mandatory Requirement}) has the highest near-tie mass (NT $=38.2\%$) and low unanimity (UY $=20.1\%$), indicating that its boundary is systematically harder to operationalize even conditional on engagement. This instability is not diffuse: disagreement concentrates in a small subset of criteria (notably \textbf{$q_4$}, \textbf{$q_5$}, \textbf{$q_9$}) rather than spreading uniformly across the schema. This concentration makes the diagnostic actionable by pinpointing specification bottlenecks.

\begin{table}[t]
\centering
\scriptsize 
\setlength{\tabcolsep}{2.2pt} 

\begin{tabular}{ll c rrr c}
\hline
\textbf{ID} & \textbf{Criterion} & \textbf{Act$_1$ (\%)} & \textbf{NT\%} & \textbf{AS\%} & \textbf{UY\%} & \textbf{$|\Omega_{q,1}|$} \\

\hline
$q_1$ & Financial Gain     & 2.8  & 22.9 & 45.8 & 31.3 & 131 \\
$q_2$ & Operational Ben.   & 9.4  & 28.2 & 43.3 & 28.4 & 443 \\
$q_3$ & Performance Imp.   & 12.6 & 28.9 & 44.8 & 26.4 & 592 \\
\hline
$q_4$ & User Well-being    & 9.7  & 35.8 & 48.3 & 15.9 & 458 \\
$q_5$ & Brand Reputation   & 17.6 & 34.0 & 52.3 & 13.7 & 826 \\
$q_6$ & Perceived Qual.    & 24.0 & 23.3 & 31.9 & 44.9 & 1130 \\
\hline
$q_7$ & Regulatory Comp.   & 5.8  & 25.6 & 44.7 & 29.7 & 273 \\
$q_8$ & Risk Mitigation    & 14.1 & 27.8 & 36.9 & 35.3 & 662 \\
$q_9$ & Mandatory Req.     & 10.3 & 38.2 & 41.7 & 20.1 & 482 \\
\hline
\end{tabular}

\caption{Criterion stability at $t=1$ ($A=5$). Act$_1(q_j)$ is activation (Eq. 3), NT (near-tie), AS (asymmetric split), and UY (unanimous yes) are derived from $\pi_j(\cdot \mid 1)$ (Eq. 4) over focused units, $|\Omega_{j,1}|$ is focus-set size. Full criterion wording: Appendix~C.}
\label{tab:criterion-stability}
\end{table}

A plausible explanation is that criteria differ in how directly they anchor to document-internal cues. Criterion \textbf{$q_6$} is often signaled by explicit evaluative markers (e.g., ``premium'', ``high-quality''), whereas \textbf{$q_4$} (\emph{User Well-being}) is frequently implicit and inference-heavy (comfort, reduced effort, health), yielding more borderline calls. This suggests that a measurable component of subjectivity arises from operationalization choices (scope, cue anchoring, examples), motivating targeted revisions to the most unstable criteria rather than blanket schema changes. These stability patterns are robust to stricter engagement thresholds and to panel perturbations (Appendix~\ref{app:robustness-rules}, Appendix~\ref{app:robustness-panel}), ensuring that we are measuring task stability rather than model bias or idiosyncratic artifacts. We next turn to structural non-identifiability: systematic overlap across categories that underdetermines single-label assignment. Model-level annotation behavior, including criterion-specific activation profiles and inter-model agreement patterns, is analyzed in 
Appendix~\ref{app:model-behavior}.





\begin{figure}[th]
  \centering
  \includegraphics[width=0.99\columnwidth]{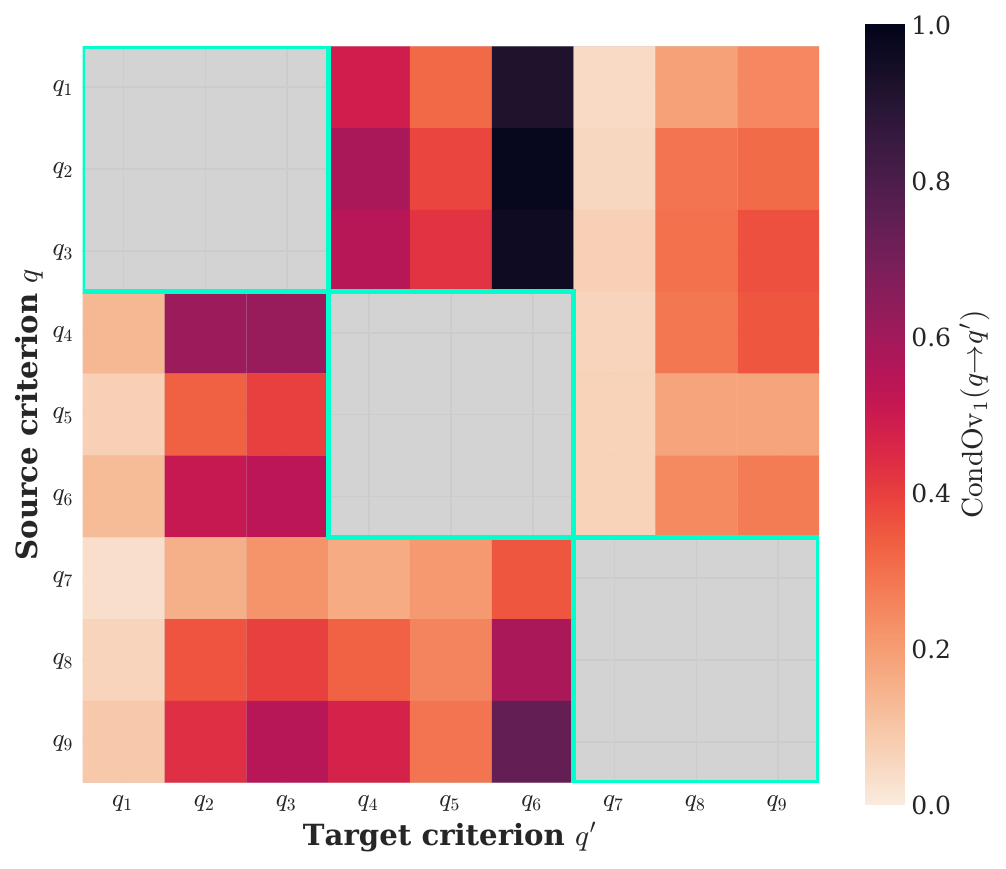}
 \caption{Cross-category leakage at $t=1$. Each cell reports directed conditional overlap
$\mathrm{CondOv}_{1}(q \!\rightarrow\! q')$, the probability that $q'$ is engaged given that $q$ is engaged.
Within-category blocks (by $\mu$) are masked to emphasize cross-category co-activation.}

  \label{fig:overlap_heatmap}
\end{figure}

\paragraph{Multi-dimensionality is measurable: overlap reveals structural bottlenecks.}
PVE is framed as mutually exclusive single-label classification, assuming persuasive evidence maps cleanly to one category. Our criterion-level audit shows that this assumption fails precisely on the subset where the schema is engaged. Table~\ref{tab:crit-count-dist} reports category overlap conditional on coverage, showing that \textbf{44.6\%} of \emph{covered} sentences activate \emph{at least two} non-target categories ($m_{s,t} \ge 2$), yielding $\mathrm{Overlap}_{\mathrm{cat} \mid \mathrm{cov},\,t} = 44.6\%$ (Eq.~\ref{eq:overlap-cat-cov}). Under a single-label design, any sentence with $m_{s,t} > 2$ is inherently underdetermined unless the guidelines provide an explicit tie-breaking policy.
In these cases the schema is \emph{non-identifying}: persuasive evidence is present ($m_{s,t} \ge 1$), yet it does not determine a unique category label. This ambiguity is already visible at the measurement layer. On covered sentences, Table~\ref{tab:crit-count-dist} shows that \textbf{64.6\%} engage at least two criteria (i.e., $|\Gamma_{s,t}| \ge 2$), and \textbf{26.5\%} engage four or more ($|\Gamma_{s,t}| \ge 4$), where $|\Gamma_{s,t}|$ is the count of criterion engagements defined in Eq.~\ref{eq:gamma}. Single-label assignment is therefore often underdetermined: annotators may agree on which criteria are present yet diverge on the final category because the schema provides no policy for resolving concurrently valid dimensions.\\
Crucially, this overlap is structured rather than diffuse. Figure~\ref{fig:overlap_heatmap} shows that a few criterion pairs dominate cross-category co-activation, localizing the strongest leakage at the $c_1$–$c_2$ boundary. These core overlap patterns are robust to stricter engagement thresholds (Appendix~\ref{app:overlap}), demonstrating they are inherent features of the schema, not methodological artifacts.
The nature of this structured overlap is revealing. Performance-related criteria often co-occur with user-experience and brand-value evidence, whereas $c_3$ (Obligation \& Safety) remains comparatively distinct. The directionality is informative: several high-leakage pairs are markedly asymmetric (as quantified by the directed conditional overlap in Eq.~\ref{eq:condov}), suggesting subset-like behavior rather than symmetric entanglement.
Together, these patterns identify a structural bottleneck in the taxonomy: commercial persuasion is inherently multi-dimensional at the $c_1$–$c_2$ boundary, so enforcing mutual exclusivity produces systematic ambiguity rather than isolated edge cases. Practically, this motivates either adding explicit tie-breaking guidance for $c_1$ vs.\ $c_2$ when both are supported, or adopting a multi-label or multi-perspective paradigm on that boundary. 
\begin{table}[htbp]
\centering
\small
\setlength{\tabcolsep}{6pt}
\renewcommand{\arraystretch}{1.1}
\begin{tabular}{ccc}
\hline
\multicolumn{3}{c}{\textbf{A. Criteria Count ($|\Gamma_{s,1}|$)}} \\
\hline
Count & \# Units & \% of covered \\
\hline
1       & 690 & 35.4 \\
2       & 419 & 21.5 \\
3       & 325 & 16.7 \\
$\ge 4$ & 517 & 26.5 \\
\hline
\multicolumn{3}{c}{\textbf{B. Category Metrics}} \\
\hline
Metric & \# Units & \% of covered \\
\hline
Covered ($m_{s,1} \ge 1$)   & 1,951 & 41.5 \\
Overlap ($m_{s,1} \ge 2$)   & 870   & 44.6 \\
Mean $|\Gamma_{s,1}|$       & 2.56  & --   \\
\hline
\end{tabular}
\caption{Measurement-layer engagement and induced category ambiguity at $t=1$.
(A) Criterion engagement counts $|\Gamma_{s,1}|$ over covered units ($m_{s,1}\!\ge\!1$), $64.7\%$ of covered units engage $\ge 2$ criteria.
(B) Coverage ($m_{s,1}\!\ge\!1$) and category overlap among covered units ($\Pr[m_{s,1}\!\ge\!2 \mid m_{s,1}\!\ge\!1]=44.6\%$).}
\label{tab:crit-count-dist}
\end{table}



\paragraph{Humans as the black box: reframing disagreement as a design signal.}
High inter-annotator disagreement on criteria such as \textbf{$q_9$} 
(``Mandatory Requirement'') is often dismissed as annotator noise or 
irreducible subjectivity. In our audit, it more often reflects a 
specification gap: the criterion underspecifies expert intent, forcing 
annotators to implicitly supply a missing decision policy (e.g., does 
``mandatory'' mean legally, contractually, or operationally required?). 
The task feels subjective not because the phenomenon is inherently ambiguous, 
but because the specification delegates intent resolution to the annotator, 
a design decision left implicit rather than made explicit.
A second, structurally distinct source of disagreement emerges even when 
criteria are individually crisp. Performance criteria (\textbf{$q_1$--$q_3$}) 
and \textbf{$q_6$} (``Perceived Quality'') frequently co-activate on sentences 
like ``This runs 10$\times$ faster and feels intuitive.'' Here, annotators 
can agree on the evidence yet disagree on the final label, because the 
single-label paradigm lacks a rule for resolving simultaneously valid 
persuasive dimensions. This is not a definitional failure but a design 
mismatch: the schema never made explicit which dimension to foreground 
when multiple are present.
Both failure modes ultimately trace back to implicit human choices in 
schema design, where more deliberate effort upfront could have prevented 
the ambiguity downstream. They demand different remedies: the first requires 
revising the \emph{instrument}, tightening scope, adding anchors, and making 
intent explicit, while the second requires revising the \emph{task design}, 
introducing tie-breaking policies, restructuring categories, or adopting 
a multi-label paradigm. The appropriate remedy also depends on the intended 
annotation paradigm. What looks like a specification gap may sometimes be 
deliberate: in a descriptive paradigm, leaving room for annotator 
interpretation is a feature, not a flaw. The diagnostic does not prescribe 
which remedy to apply; it makes the source of disagreement explicit and 
reportable, turning disagreement from an endpoint into a starting point 
for principled revision. Concrete examples of how these signals guided 
criterion-level refinements are discussed in Appendix~\ref{app:refinement}.


\paragraph{Human validation: diagnostic signals predict and explain expert disagreement.}
To validate the diagnostic, we test whether its signals align with expert judgment on a held-out set of 500 sentences labeled by five domain experts. The correspondence is strong. Sentences flagged as cross-category co-activated ($m_{s,t}\ge 2$) show substantially higher expert disagreement than single-category cases ($m_{s,t}=1$). Critically, the diagnostic pinpoints the same problematic boundary that most confuses experts: the $c_1$--$c_2$ boundary shows both the highest diagnostic co-activation (83.8\%) and the highest expert split rate (37.5\%), far above other boundaries. This alignment confirms that the diagnostic reveals stable, schema-level flaws, not artifacts of the LLM panel. More importantly, it provides an actionable intervention map. We can now distinguish whether expert struggle stems from \emph{criterion ambiguity} (e.g., underspecified definitions) or a \emph{schema mismatch} (multi-dimensional evidence forced into a single label). Each requires a distinct remedy, tightening definitions versus adding tie-breaking rules or shifting paradigms, as illustrated qualitatively in Appendix~\ref{app:qual-examples}. Thus, the diagnostic completes the diagnosis phase, providing an evidence-based map for targeted intervention. This transforms schema refinement from guesswork into a precise engineering task: first diagnose the source of disagreement, then apply the appropriate remedy.

\begin{table}[t]
\centering
\small
\setlength{\tabcolsep}{8pt}
\renewcommand{\arraystretch}{1.1}
\begin{tabular}{ccc}
\hline
\textbf{Pair} & \textbf{Human split (\%)} & \textbf{Diag.\ co-act (\%)} \\
\hline
$c_1$--$c_2$ & 37.5 & 83.8 \\
$c_1$--$c_3$ & 16.5 & 44.4 \\
$c_2$--$c_3$ & 14.9 & 50.7 \\
\hline
\end{tabular}
\caption{Boundary alignment on the human-validation subset (covered sentences only at $t=1$, $m_{s,1}\ge 1$).
\textbf{Human split}: fraction of covered sentences where experts assign both $c_a$ and $c_b$.
\textbf{Diag.\ co-act}: fraction of covered sentences where both categories are diagnostically active (Eq.~\ref{eq:gik}).}
\label{tab:human-validation}
\end{table}


\section{Conclusion}
Progress in subjective NLP requires moving beyond treating annotator disagreement as mere noise. We introduce a diagnostic that instead characterizes its source, distinguishing between instability from underspecified criteria and non-separability from intrinsically multidimensional evidence.
This characterization provides a critical, evidence-based map for schema design. It allows designers to make a principled choice: to operationalize a prescriptive task by tightening definitions and adding tie-breakers, or to formalize a descriptive task by adopting a multi-label or multi-perspective paradigm that captures legitimate plurality.
Thus, our work shifts the goal from enforcing consensus to understanding the task structure. By making these design choices explicit and auditable, we provide a foundation to reorient benchmark creation for building resources that are not just larger, but more interpretable, deliberate, and authentically aligned with the phenomena they aim to capture.

\newpage\section{Limitations}
\label{sec:limitations}

\paragraph{Limitation: LLM-based annotation panel.}
Our diagnostic relies on an LLM panel to enable scalable and reproducible schema stress-testing, a pragmatic choice for auditing written criteria. The trade-off is that LLM judgments may reflect shared training priors or similar instruction-following behavior, which can shape the observed patterns of instability and overlap.
Our held-out expert validation indicates that the main diagnostic signals are meaningful, but LLM-specific artifacts may persist, especially for niche or domain-specific criteria. Accordingly, these results should be interpreted as revealing the schema’s \emph{failure modes under a consistent automated judgment regime}, not as estimates of population-level human agreement. Full validation requires replication with diverse human panels and hybrid human–LLM designs to characterize where judgments converge and where they systematically diverge.

\paragraph{Corpus specificity and generalization.}
Our analysis is grounded in one corpus of real commercial documents. While this 
improves ecological validity for PVE, diagnostics are corpus-sensitive: different 
organizations, sectors, or preprocessing choices may shift which criteria appear 
unstable or entangled. The diagnostic is most informative when run on the actual 
target distribution and repeated across diverse corpora.

\paragraph{Single-task instantiation.}
We validate the diagnostic on one schema and one domain: Persuasive Value Extraction in commercial documents. While the diagnostic framework is model-agnostic and applies to any task with human-defined criteria and multi-annotator criterion judgments, empirical conclusions about which failure modes dominate and how overlap concentrates may not transfer to other domains, label spaces, or discourse genres. Demonstrating generality requires replication on additional subjective tasks with independently designed schemas and different operational constraints (e.g., moderation, stance, clinical narratives).

\paragraph{Criterion wording, binarization, and language dependence.}
Our criteria are binary yes/no questions under an explicit-only rule, and diagnostic outcomes inherit these design choices. Binarization compresses graded evidence and can increase near-ties around implicit thresholds. Moreover, the corpus is primarily French with English technical terminology; lexical cues, modality markers, and regulatory language vary significantly across languages and sectors. As a result, stability and overlap should be interpreted as properties of this operationalization in this linguistic setting, not as language-invariant properties of the underlying constructs. Applying the diagnostic elsewhere will require re-calibration of criterion wording and may change which boundaries appear unstable or entangled.

\paragraph{Task framing and the single-label constraint.}
PVE is framed as single-label due to downstream procurement workflows, yet the corpus frequently expresses multiple value dimensions in the same sentence. Observed co-activation may therefore reflect a mismatch between the phenomenon and the forced-choice framing as much as any schema deficiency. Our diagnostic localizes where this mismatch is concentrated, but it does not determine which resolution is appropriate (tighten boundaries, restructure categories, adopt multi-label, or impose a stakeholder-specific tie-breaker). These choices depend on external requirements and normative commitments outside the diagnostic.

\paragraph{Data access and reproducibility.}
Raw source documents cannot be released due to client confidentiality. 
The released repository contains anonymized sentence data and the full 
annotation tensor, enabling reproduction of all diagnostic procedures. 
End-to-end reproducibility from raw documents is not possible, but 
all diagnostic methodology is fully specified and applicable to other datasets.


\paragraph{Scope and future work.}
Our diagnostic flags unstable or non-separable criteria but does not prescribe solutions. Future work should validate on public datasets with diverse annotator pools, connect diagnostic signals to concrete interventions (revised definitions, multi-label protocols, tie-break policies), and evaluate downstream effects on evaluation across domains and cultures. This aligns with emerging shared-task efforts replacing coarse labels with dimensional supervision (e.g., SemEval-2026 Task 3 Track B: DimStance).


\newpage
\bibliography{custom}

@book{chochoy2025msmkc,
  author    = {Chochoy, Emmanuel},
  title     = {La m{\'e}thode {MSMKC} : R{\'e}inventer la vente {\`a} l'{\`e}re du chaos num{\'e}rique},
  year      = {2025},
  publisher = {Economica \& L'{\'E}diteur {\`a} part},
  isbn      = {978-2-494780-19-4},
}

@article{aroyo_truth_2015,
	title = {Truth {Is} a {Lie}: {Crowd} {Truth} and the {Seven} {Myths} of {Human} {Annotation}},
	volume = {36},
	copyright = {http://onlinelibrary.wiley.com/termsAndConditions\#vor},
	issn = {0738-4602, 2371-9621},
	shorttitle = {Truth {Is} a {Lie}},
	url = {https://onlinelibrary.wiley.com/doi/10.1609/aimag.v36i1.2564},
	doi = {10.1609/aimag.v36i1.2564},
	language = {en},
	number = {1},
	urldate = {2025-11-27},
	journal = {AI Magazine},
	author = {Aroyo, Lora and Welty, Chris},
	month = mar,
	year = {2015},
	pages = {15--24}
}

@article{pavlick_inherent_2019,
	title = {Inherent {Disagreements} in {Human} {Textual} {Inferences}},
	volume = {7},
	url = {https://aclanthology.org/Q19-1043/},
	doi = {10.1162/tacl_a_00293},
	urldate = {2025-11-27},
	journal = {Transactions of the Association for Computational Linguistics},
	author = {Pavlick, Ellie and Kwiatkowski, Tom},
	editor = {Lee, Lillian and Johnson, Mark and Roark, Brian and Nenkova, Ani},
	year = {2019},
	note = {Place: Cambridge, MA
Publisher: MIT Press},
	pages = {677--694}
}

@inproceedings{plank_problem_2022,
	address = {Abu Dhabi, United Arab Emirates},
	title = {The “{Problem}” of {Human} {Label} {Variation}: {On} {Ground} {Truth} in {Data}, {Modeling} and {Evaluation}},
	shorttitle = {The “{Problem}” of {Human} {Label} {Variation}},
	url = {https://aclanthology.org/2022.emnlp-main.731/},
	doi = {10.18653/v1/2022.emnlp-main.731},
	urldate = {2025-11-27},
	booktitle = {Proceedings of the 2022 {Conference} on {Empirical} {Methods} in {Natural} {Language} {Processing}},
	publisher = {Association for Computational Linguistics},
	author = {Plank, Barbara},
	editor = {Goldberg, Yoav and Kozareva, Zornitsa and Zhang, Yue},
	month = dec,
	year = {2022},
	pages = {10671--10682}
}

@article{uma_learning_2021,
	title = {Learning from {Disagreement}: {A} {Survey}},
	volume = {72},
	copyright = {Copyright (c)},
	issn = {1076-9757},
	shorttitle = {Learning from {Disagreement}},
	url = {https://jair.org/index.php/jair/article/view/12752},
	doi = {10.1613/jair.1.12752},
	language = {en},
	urldate = {2025-11-27},
	journal = {Journal of Artificial Intelligence Research},
	author = {Uma, Alexandra N. and Fornaciari, Tommaso and Hovy, Dirk and Paun, Silviu and Plank, Barbara and Poesio, Massimo},
	month = dec,
	year = {2021},
	pages = {1385--1470}
}

@article{mostafazadeh_davani_dealing_2022,
	title = {Dealing with {Disagreements}: {Looking} {Beyond} the {Majority} {Vote} in {Subjective} {Annotations}},
	volume = {10},
	shorttitle = {Dealing with {Disagreements}},
	url = {https://aclanthology.org/2022.tacl-1.6/},
	doi = {10.1162/tacl_a_00449},
	urldate = {2025-11-27},
	journal = {Transactions of the Association for Computational Linguistics},
	author = {Mostafazadeh Davani, Aida and Díaz, Mark and Prabhakaran, Vinodkumar},
	editor = {Roark, Brian and Nenkova, Ani},
	year = {2022},
	note = {Place: Cambridge, MA
Publisher: MIT Press},
	pages = {92--110},

}

@inproceedings{swayamdipta_dataset_2020,
	address = {Online},
	title = {Dataset {Cartography}: {Mapping} and {Diagnosing} {Datasets} with {Training} {Dynamics}},
	shorttitle = {Dataset {Cartography}},
	url = {https://aclanthology.org/2020.emnlp-main.746/},
	doi = {10.18653/v1/2020.emnlp-main.746},
	urldate = {2025-12-06},
	booktitle = {Proceedings of the 2020 {Conference} on {Empirical} {Methods} in {Natural} {Language} {Processing} ({EMNLP})},
	publisher = {Association for Computational Linguistics},
	author = {Swayamdipta, Swabha and Schwartz, Roy and Lourie, Nicholas and Wang, Yizhong and Hajishirzi, Hannaneh and Smith, Noah A. and Choi, Yejin},
	editor = {Webber, Bonnie and Cohn, Trevor and He, Yulan and Liu, Yang},
	month = nov,
	year = {2020},
	pages = {9275--9293}
}

@article{basile_toward_2023,
	title = {Toward a {Perspectivist} {Turn} in {Ground} {Truthing} for {Predictive} {Computing}},
	volume = {37},
	issn = {2374-3468, 2159-5399},
	url = {http://arxiv.org/abs/2109.04270},
	doi = {10.1609/aaai.v37i6.25840},
	number = {6},
	urldate = {2025-12-03},
	journal = {Proceedings of the AAAI Conference on Artificial Intelligence},
	author = {Basile, Valerio and Cabitza, Federico and Campagner, Andrea and Fell, Michael},
	month = jun,
	year = {2023},
	note = {arXiv:2109.04270 [cs]},
	pages = {6860--6868}
}

@inproceedings{sap_social_2020,
	address = {Online},
	title = {Social {Bias} {Frames}: {Reasoning} about {Social} and {Power} {Implications} of {Language}},
	shorttitle = {Social {Bias} {Frames}},
	url = {https://aclanthology.org/2020.acl-main.486/},
	doi = {10.18653/v1/2020.acl-main.486},
	urldate = {2025-11-27},
	booktitle = {Proceedings of the 58th {Annual} {Meeting} of the {Association} for {Computational} {Linguistics}},
	publisher = {Association for Computational Linguistics},
	author = {Sap, Maarten and Gabriel, Saadia and Qin, Lianhui and Jurafsky, Dan and Smith, Noah A. and Choi, Yejin},
	editor = {Jurafsky, Dan and Chai, Joyce and Schluter, Natalie and Tetreault, Joel},
	month = jul,
	year = {2020},
	pages = {5477--5490}
}

@misc{ruggeri_let_2024,
	title = {Let {Guidelines} {Guide} {You}: {A} {Prescriptive} {Guideline}-{Centered} {Data} {Annotation} {Methodology}},
	shorttitle = {Let {Guidelines} {Guide} {You}},
	url = {http://arxiv.org/abs/2406.14099},
	doi = {10.48550/arXiv.2406.14099},
	urldate = {2025-12-02},
	publisher = {arXiv},
	author = {Ruggeri, Federico and Misino, Eleonora and Muti, Arianna and Korre, Katerina and Torroni, Paolo and Barrón-Cedeño, Alberto},
	month = dec,
	year = {2024},
	note = {arXiv:2406.14099 [cs]}
}

@misc{jikeli_antisemitic_2023,
	title = {Antisemitic {Messages}? {A} {Guide} to {High}-{Quality} {Annotation} and a {Labeled} {Dataset} of {Tweets}},
	shorttitle = {Antisemitic {Messages}?},
	url = {http://arxiv.org/abs/2304.14599},
	doi = {10.48550/arXiv.2304.14599},
	urldate = {2025-12-03},
	publisher = {arXiv},
	author = {Jikeli, Gunther and Karali, Sameer and Miehling, Daniel and Soemer, Katharina},
	month = apr,
	year = {2023},
	note = {arXiv:2304.14599 [cs]}
}

@inproceedings{rottger_two_2022,
	address = {Seattle, United States},
	title = {Two {Contrasting} {Data} {Annotation} {Paradigms} for {Subjective} {NLP} {Tasks}},
	url = {https://aclanthology.org/2022.naacl-main.13/},
	doi = {10.18653/v1/2022.naacl-main.13},
	urldate = {2025-11-27},
	booktitle = {Proceedings of the 2022 {Conference} of the {North} {American} {Chapter} of the {Association} for {Computational} {Linguistics}: {Human} {Language} {Technologies}},
	publisher = {Association for Computational Linguistics},
	author = {Röttger, Paul and Vidgen, Bertie and Hovy, Dirk and Pierrehumbert, Janet},
	editor = {Carpuat, Marine and de Marneffe, Marie-Catherine and Meza Ruiz, Ivan Vladimir},
	month = jul,
	year = {2022},
	pages = {175--190}
}

@misc{northcutt_pervasive_2021,
	title = {Pervasive {Label} {Errors} in {Test} {Sets} {Destabilize} {Machine} {Learning} {Benchmarks}},
	url = {http://arxiv.org/abs/2103.14749},
	doi = {10.48550/arXiv.2103.14749},
	urldate = {2025-12-09},
	publisher = {arXiv},
	author = {Northcutt, Curtis G. and Athalye, Anish and Mueller, Jonas},
	month = nov,
	year = {2021},
	note = {arXiv:2103.14749 [stat]},
}

@misc{korre_untangling_2025,
	title = {Untangling {Hate} {Speech} {Definitions}: {A} {Semantic} {Componential} {Analysis} {Across} {Cultures} and {Domains}},
	shorttitle = {Untangling {Hate} {Speech} {Definitions}},
	url = {http://arxiv.org/abs/2411.07417},
	doi = {10.48550/arXiv.2411.07417},
	urldate = {2025-12-10},
	publisher = {arXiv},
	author = {Korre, Katerina and Muti, Arianna and Ruggeri, Federico and Barrón-Cedeño, Alberto},
	month = may,
	year = {2025},
	note = {arXiv:2411.07417 [cs]}
}

@article{salminen_anatomy_2018,
	title = {Anatomy of {Online} {Hate}: {Developing} a {Taxonomy} and {Machine} {Learning} {Models} for {Identifying} and {Classifying} {Hate} in {Online} {News} {Media}},
	volume = {12},
	copyright = {Copyright (c) 2022 Proceedings of the International AAAI Conference on Web and Social Media},
	issn = {2334-0770},
	shorttitle = {Anatomy of {Online} {Hate}},
	url = {https://ojs.aaai.org/index.php/ICWSM/article/view/15028},
	doi = {10.1609/icwsm.v12i1.15028},
	language = {en},
	number = {1},
	urldate = {2025-12-10},
	journal = {Proceedings of the International AAAI Conference on Web and Social Media},
	author = {Salminen, Joni and Almerekhi, Hind and Milenković, Milica and Jung, Soon-gyo and An, Jisun and Kwak, Haewoon and Jansen, Bernard},
	month = jun,
	year = {2018}
}

@misc{ignatev_hypernetworks_2025,
	title = {Hypernetworks for {Perspectivist} {Adaptation}},
	url = {http://arxiv.org/abs/2510.13259},
	doi = {10.48550/arXiv.2510.13259},
	urldate = {2025-12-10},
	publisher = {arXiv},
	author = {Ignatev, Daniil and Paperno, Denis and Poesio, Massimo},
	month = oct,
	year = {2025},
	note = {arXiv:2510.13259 [cs]}
}

@article{artstein_survey_2008,
	title = {Survey {Article}: {Inter}-{Coder} {Agreement} for {Computational} {Linguistics}},
	volume = {34},
	shorttitle = {Survey {Article}},
	url = {https://aclanthology.org/J08-4004/},
	doi = {10.1162/coli.07-034-R2},
	number = {4},
	urldate = {2025-12-10},
	journal = {Computational Linguistics},
	author = {Artstein, Ron and Poesio, Massimo},
	year = {2008},
	pages = {555--596}
}

@article{paun_comparing_2018,
	title = {Comparing {Bayesian} {Models} of {Annotation}},
	volume = {6},
	url = {https://aclanthology.org/Q18-1040/},
	doi = {10.1162/tacl_a_00040},
	urldate = {2025-12-10},
	journal = {Transactions of the Association for Computational Linguistics},
	author = {Paun, Silviu and Carpenter, Bob and Chamberlain, Jon and Hovy, Dirk and Kruschwitz, Udo and Poesio, Massimo},
	editor = {Lee, Lillian and Johnson, Mark and Toutanova, Kristina and Roark, Brian},
	year = {2018},
	note = {Place: Cambridge, MA
Publisher: MIT Press},
	pages = {571--585}
}

@inproceedings{plepi_unifying_2022,
	address = {Abu Dhabi, United Arab Emirates},
	title = {Unifying {Data} {Perspectivism} and {Personalization}: {An} {Application} to {Social} {Norms}},
	shorttitle = {Unifying {Data} {Perspectivism} and {Personalization}},
	url = {https://aclanthology.org/2022.emnlp-main.500/},
	doi = {10.18653/v1/2022.emnlp-main.500},
	urldate = {2025-12-10},
	booktitle = {Proceedings of the 2022 {Conference} on {Empirical} {Methods} in {Natural} {Language} {Processing}},
	publisher = {Association for Computational Linguistics},
	author = {Plepi, Joan and Neuendorf, Béla and Flek, Lucie and Welch, Charles},
	editor = {Goldberg, Yoav and Kozareva, Zornitsa and Zhang, Yue},
	month = dec,
	year = {2022},
	pages = {7391--7402}
	}

@article{bender_data_2018,
	title = {Data {Statements} for {Natural} {Language} {Processing}: {Toward} {Mitigating} {System} {Bias} and {Enabling} {Better} {Science}},
	volume = {6},
	shorttitle = {Data {Statements} for {Natural} {Language} {Processing}},
	url = {https://aclanthology.org/Q18-1041/},
	doi = {10.1162/tacl_a_00041},
	urldate = {2025-12-10},
	journal = {Transactions of the Association for Computational Linguistics},
	author = {Bender, Emily M. and Friedman, Batya},
	editor = {Lee, Lillian and Johnson, Mark and Toutanova, Kristina and Roark, Brian},
	year = {2018},
	note = {Place: Cambridge, MA
Publisher: MIT Press},
	pages = {587--604}
}

@misc{gebru_datasheets_2021,
	title = {Datasheets for {Datasets}},
	url = {http://arxiv.org/abs/1803.09010},
	doi = {10.48550/arXiv.1803.09010},
	urldate = {2025-12-10},
	publisher = {arXiv},
	author = {Gebru, Timnit and Morgenstern, Jamie and Vecchione, Briana and Vaughan, Jennifer Wortman and Wallach, Hanna and Daumé, Hal and Crawford, Kate},
	month = dec,
	year = {2021},
	note = {arXiv:1803.09010 [cs]}
}

@article{dawid_maximum_1979,
	title = {Maximum {Likelihood} {Estimation} of {Observer} {Error}-{Rates} {Using} the {EM} {Algorithm}},
	volume = {28},
	issn = {0035-9254},
	url = {https://www.jstor.org/stable/2346806},
	doi = {10.2307/2346806},
	number = {1},
	urldate = {2025-12-10},
	journal = {Journal of the Royal Statistical Society. Series C (Applied Statistics)},
	author = {Dawid, A. P. and Skene, A. M.},
	year = {1979},
	note = {Publisher: [Royal Statistical Society, Oxford University Press]},
	pages = {20--28}
}

@inproceedings{basile_we_2021,
	address = {Online},
	title = {We {Need} to {Consider} {Disagreement} in {Evaluation}},
	url = {https://aclanthology.org/2021.bppf-1.3/},
	doi = {10.18653/v1/2021.bppf-1.3},
	urldate = {2025-12-11},
	booktitle = {Proceedings of the 1st {Workshop} on {Benchmarking}: {Past}, {Present} and {Future}},
	publisher = {Association for Computational Linguistics},
	author = {Basile, Valerio and Fell, Michael and Fornaciari, Tommaso and Hovy, Dirk and Paun, Silviu and Plank, Barbara and Poesio, Massimo and Uma, Alexandra},
	editor = {Church, Kenneth and Liberman, Mark and Kordoni, Valia},
	month = aug,
	year = {2021},
	pages = {15--21}
}
\clearpage
\onecolumn
\appendix

\section{Supplementary Method Details}
\label{app:notation}

\paragraph{Notation consistency.} 
We use the same indices as in Section~\ref{sec:method}: $s$ indexes sentences (units), $q$ indexes criteria, and $a$ indexes annotators.
Table~\ref{tab:notation} summarizes the notation used in Section~\ref{sec:method} for quick reference.
\begin{table}[th]
\centering
\small
\begin{tabular}{p{0.20\linewidth} p{0.7\linewidth}}
\hline
\textbf{Symbol} & \textbf{Meaning} \\
\hline
$\mathcal{T}=(\mathcal{C},\mathcal{Q},\mu)$
& Task schema: categories $\mathcal{C}$, criteria $\mathcal{Q}$, and mapping $\mu$. \\

$\mathcal{C}=\{c_0,c_1,\dots,c_{C}\}$
& Category set ($|\mathcal{C}|=C+1$); $c_0$ denotes the non-target / default category. \\

$\mathcal{Q}=\{q_1,\dots,q_Q\}$
& Set of expert-defined binary criteria (yes/no questions), $|\mathcal{Q}|=Q$. \\

$\mu:\mathcal{Q}\rightarrow \mathcal{C}$
& Mapping from each criterion to the (single) category it is intended to support. \\

$\mathcal{Q}_c$
& Supporting criteria for category $c_c$:
$\mathcal{Q}_c=\{\,q\in\mathcal{Q}\mid \mu(q)=c_c\,\}$. \\

$\mathcal{S}=\{s_1,\dots,s_N\}$
& Corpus of annotation units (e.g., sentences), with $|\mathcal{S}|=N$. \\

$\mathcal{A}=\{a_1,\dots,a_A\}$
& Annotator panel of size $A$. \\

$\mathbf{Y}\in\{0,1\}^{|\mathcal{S}|\times A\times Q}$
& Binary response tensor; $y_{saq}=1$ if annotator $a$ marks criterion $q$ as present in unit $s$. \\

$v_{sq}$
& Positive vote count for $(s,q)$:
$v_{sq}=\sum_{a=1}^{A} y_{saq}\in\{0,\dots,A\}$. \\

$t$
& Vote threshold used to define focus sets, $t\in\{0,1,\dots,A\}$. \\

$\Omega_{q,t}$
& Focus set for criterion $q$ at threshold $t$:
$\Omega_{q,t}=\{\,s\in\mathcal{S}\mid v_{sq}\ge t\,\}$. \\

$\mathrm{Act}_{t}(q)$
& Activation rate at threshold $t$:
$\mathrm{Act}_{t}(q)=\frac{|\Omega_{q,t}|}{|\mathcal{S}|}$. \\

$\pi_q(k\mid t)$
& Conditional vote distribution over $k\in\{t,\dots,A\}$:
$\pi_q(k\mid t)=\frac{|\{\, s\in\Omega_{q,t}\mid v_{sq}=k \,\}|}{|\Omega_{q,t}|}$. \\

$\Gamma_{s,t}$
& Engaged criteria for unit $s$ at threshold $t$:
$\Gamma_{s,t}=\{\,q\in\mathcal{Q}\mid s\in\Omega_{q,t}\,\}$. \\

$\mathrm{Overlap}_{\mathrm{crit},t}$
& Criterion-level overlap rate at threshold $t$:
$\mathrm{Overlap}_{\mathrm{crit},t}
=\frac{1}{|\mathcal{S}|}\sum_{s\in\mathcal{S}}{\bf 1}\!\left[|\Gamma_{s,t}|\ge 2\right]$. \\

$\mathrm{CondOv}_{t}(q\rightarrow q')$
& Directed conditional overlap at threshold $t$:
$\mathrm{CondOv}_{t}(q\rightarrow q')
=\frac{|\Omega_{q,t}\cap\Omega_{q',t}|}{|\Omega_{q,t}|}$ (asymmetric \hspace{0pt} in $q,q'$). \\

$g_{sc,t}$
& Category engagement indicator at threshold $t$:
$g_{sc,t}={\bf 1}\!\left[\exists\, q\in\mathcal{Q}_c:\ s\in\Omega_{q,t}\right]$. \\

$m_{s,t}$
& Number of active non-target categories for unit $s$ at threshold $t$:
$m_{s,t}=\sum_{c=1}^{C} g_{sc,t}$. \\

$\mathrm{Overlap}_{\mathrm{cat},t}$
& Category-level overlap rate at threshold $t$:
$\mathrm{Overlap}_{\mathrm{cat},t}
=\frac{1}{|\mathcal{S}|}\sum_{s\in\mathcal{S}}{\bf 1}\!\left[m_{s,t}\ge 2\right]$. \\

$\mathrm{Overlap}_{\mathrm{cat}\mid \mathrm{cov},t}$
& Category overlap conditional on coverage ($m_{s,t}\ge 1$):
$\mathrm{Overlap}_{\mathrm{cat}\mid \mathrm{cov},t}
=\frac{\sum_{s\in\mathcal{S}} {\bf 1}\!\left[m_{s,t}\ge2\right]}
{\sum_{s\in\mathcal{S}} {\bf 1}\!\left[m_{s,t}\ge1\right]}$. \\
\hline
\end{tabular}
\caption{Notation used in Section~\ref{sec:method}.}
\label{tab:notation}
\end{table}

\section{Task Examples and Boundary Cases}
\label{app:task-examples}
        
    \paragraph{Why PVE matters in practice.}
    PVE arises from a commercial document-analysis workflow used by a procurement-focused consulting partner.
    In this setting, analysts read vendor-facing documents (offers, technical notes, product brochures) and restructure them into decision-ready summaries for buyers.
    A recurring pain point is that persuasive content is interleaved with technical description, boilerplate, and formatting artefacts, making it time-consuming to identify \emph{what actually justifies the purchase}.
    PVE therefore targets sentences that express a procurement-relevant justification, rather than descriptive background.
            
    \paragraph{Practitioner rationale for the four categories.}
    The schema is grounded in a practitioner view that buyers typically justify purchases along a small set of recurring axes.
    In our partner's framework, a sentence is persuasive when it supports at least one of four broad motivations:
    (i) improved performance or efficiency (e.g., productivity, savings, measurable outcomes),
    (ii) improved experience or brand value (e.g., perceived quality, attractiveness, reputation, trust),
    (iii) obligation or safety (e.g., compliance requirements, risk reduction, security), or
    (iv) none of the above (descriptive content that does not provide a justification).
    These axes reflect how procurement teams communicate internally: they map technical statements to decision narratives that are financial, experiential/reputational, or compliance- and risk-driven.
            
    \paragraph{What makes PVE difficult.}
    Commercial writing often compresses multiple motivations into one sentence (e.g., automation \emph{and} compliance, security \emph{and} trust), and the same statement can be plausibly interpreted through different organizational priorities.
    This produces boundary cases under a single-label design: even when all annotators act in good faith, choosing one label can require committing to one foregrounded rationale and down-weighting others.\\
    A further source of difficulty is that persuasive value is often implicit 
    rather than explicit: the same statement can carry different weight depending 
    on who is reading it, their organizational role, and the procurement context. 
    Annotators with domain expertise naturally draw on this background knowledge, 
    sometimes completing the interpretation beyond what is explicitly stated, 
    introducing variation that reflects legitimate contextual reasoning rather 
    than annotation error.
                    
    \paragraph{Illustrative examples.}
    Table~\ref{tab:task-examples} provides representative cases spanning clear non-activations, strong multi-criterion activations, and boundary cases with LLM disagreement.
    These examples illustrate instability (split votes) and overlap (multiple criteria activating simultaneously), the core phenomena detected by our diagnostic.
            
    \begin{table}[th]
    \centering
    \small
    \setlength{\tabcolsep}{5pt}
    \renewcommand{\arraystretch}{1.25}
    \resizebox{\textwidth}{!}{
    \begin{tabular}{p{1.0cm} p{11.7cm} p{6.3cm}}
    \hline
    \textbf{ID} &
    \textbf{Sentence (English)} &
    \textbf{LLM criterion votes (Yes/No, panel size $=5$)} \\
    \hline
    
    2414 &
    In this context and at the request of elected officials, a regional agricultural authority offers a new seasonal accommodation service. &
    None triggered (all 0/5) \\
    
    233 &
    Ecological recycling of end-of-life IT equipment through awareness campaigns, sorting points, collection of used materials, secure document destruction and GDPR-compliant data erasure, followed by refurbishment of reusable components. &
    $q_5$:(2/3), $q_6$:(1/4), $q_7$:(5/0), $q_8$:(5/0), $q_9$:(3/2) \\
    
    887 &
    We help reduce employees' commute times to improve quality of life and reduce lateness, absenteeism, turnover, and carbon footprint. &
    $q_2$:(5/0), $q_3$:(5/0), $q_4$:(5/0), $q_6$:(5/0) \\
    
    159 &
    Consider filing an export-support application to obtain a subsidy for international expansion. &
    $q_1$:(3/2) \\
    
    708 &
    Our agreement specifies GDPR-compliant data processing procedures between you, our organization, and the employer providing the data. &
    $q_5$:(1/4), $q_6$:(2/3), $q_7$:(5/0), $q_8$:(5/0), $q_9$:(4/1) \\
    
    2360 &
    To ensure continuity, we must give young professionals visibility through profitability, projects, and recognition via a positive public image. &
    $q_1$:(4/1), $q_3$:(2/3), $q_4$:(2/3), $q_5$:(5/0), $q_6$:(5/0), $q_9$:(4/1) \\
    
    4555 &
    Our innovation protects tree bases against erosion and compaction, enabling better rooting and improved infiltration of runoff water. &
    $q_2$:(5/0), $q_3$:(4/1), $q_4$:(1/4), $q_6$:(5/0), $q_8$:(5/0), $q_9$:(3/2) \\
    
    \hline
    \end{tabular}
    }
    \caption{Representative examples illustrating criterion activation patterns at $t=1$. Vote tallies (Yes/No out of 5 LLM models) reveal consistent activations (e.g., 5/0), borderline splits (e.g., 3/2), and non-activations. These patterns illustrate the instability and overlap phenomena detected by our diagnostic.}
    \label{tab:task-examples}
    \end{table}

     \section{Persuasive Value Extraction Criteria}
\label{app:criteria}

\paragraph{From coarse labels to operational criteria.}
When we first discussed persuasive value with our procurement-focused consulting partner, they could readily distinguish non-persuasive content (technical description, boilerplate, formatting artefacts) from persuasive content (sentences that justify a purchase decision). The difficulty emerged when translating this intuition into a stable annotation scheme: even practitioners disagreed on why a sentence is persuasive, who benefits from the stated action, and whether the value is directly stated or only inferred. Rather than treating this as a fixed taxonomy from the start, the definition evolved through calibration discussions, and experts converged on nine binary criteria (\(q_1\)--\(q_9\)), phrased as yes/no questions, to capture recurring procurement rationales and make disagreements observable rather than hidden under a single label.
\paragraph{Iterative calibration process.}
The nine criteria were initially drafted by the lead coordinator based on 
domain analysis of persuasive value in B2B commercial documents. These draft 
constructs were then discussed and refined in structured workshops with domain 
experts prior to any large-scale annotation. The process involved four stages: 
(i) construct decomposition into binary decision rules, (ii) pilot annotation 
on a subset of documents, (iii) structured discussion of disagreement cases 
supported by concrete sentence examples reviewed collectively, and (iv) wording 
refinement guided by observed boundary failures. Criteria definitions are treated 
as living artifacts: they remain open to revision as new boundary cases surface, 
and the diagnostic is precisely designed to support this ongoing stabilization.

A central challenge during calibration was separating signals that are 
conceptually related but operationally distinct. For instance, defining 
what counts as a return on investment required explicit choices: does ROI 
refer strictly to measurable financial return, or does any stated gain 
qualify? Such decisions were made collaboratively by domain experts and 
reflect the company's current operational priorities rather than universal 
definitional truths. These choices are assumed for now but remain open to 
revision as the schema matures. Once sufficiently stable, the schema could 
itself become an object of study, for instance to investigate how 
procurement-relevant value is constructed and communicated in commercial 
discourse, at the intersection of organizational sociology and computational 
linguistics.

To illustrate, the initial formulation of $q_5$ (Reputation \& Recognition) 
conflated brand visibility signals with perceived quality judgments, a tension 
that only became measurable after applying the diagnostic. This signal prompted 
structured reconsideration, leading to the introduction of a refined criterion 
and reduced overlap between recognition and quality signals. This exemplifies 
how the diagnostic supports iterative schema development rather than prescribing 
a fixed endpoint.

For practitioners applying this framework to new tasks, we recommend: 
(i) explicit construct decomposition, (ii) operationalizing each construct 
as clear binary decision rules, (iii) pilot annotation with structured 
disagreement review supported by shared sentence examples, 
(iv) explicit articulation of the chosen annotation paradigm prior to 
large-scale annotation, and (v) diagnostic evaluation prior to committing 
to gold labels.
\paragraph{Why an explicit-only rule was adopted.}
One key design choice to emerge from calibration was the adoption of an 
explicit-only annotation rule, motivated by the need to anchor judgments 
to textual evidence rather than background assumptions. During calibration, experts repeatedly encountered sentences that support multiple plausible readings once background assumptions are introduced. For example, a capability might seem to save time, or a policy change might produce financial return, but these inferences require stepping beyond what the text explicitly states. This created unstable judgments, especially when distinguishing between performance/efficiency rationales and experience/brand-value rationales, because annotators could legitimately ``complete the story'' beyond the sentence's explicit content. Consider the sentence: ``We are deploying more police officers in the city.'' Different experts foreground different rationales: one sees this as a compliance/safety measure (public security mandate), another as improved user well-being (citizens feel safer), while a third connects it indirectly to economic value (a safer city attracts tourism and revenue).

To reduce this drift and keep the criteria anchored to textual evidence, experts adopted an explicit-only principle: a criterion is marked \texttt{Yes} only if the corresponding value claim is explicitly stated in the sentence. If the value requires inference or external context, the answer should be \texttt{No}. Each criterion is evaluated independently as \texttt{Yes}/\texttt{No}.

\paragraph{Examples of the explicit-only rule in practice.}
Below are illustrative cases showing how the explicit-only rule prevents ``story completion'' beyond the sentence:

\begin{itemize}
  \setlength{\itemsep}{2pt}

  \item \textbf{Inference-only capability (explicit-only: \texttt{No}).}
  ``The platform provides automated incident workflows and configurable dashboards.''
  \emph{Tempting inference:} time savings / productivity.
  \emph{Explicit-only:} \texttt{No} for \(q_2\) (Operational Efficiency) and \(q_3\) (Organizational Impact), since no benefit (faster, reduced workload, improved results) is explicitly stated.

  \item \textbf{Explicit operational gain (explicit-only: \texttt{Yes}).}
  ``Automation reduces processing time by 30\% and cuts manual workload.''
  \emph{Explicit-only:} \texttt{Yes} for \(q_2\) (Operational Efficiency) because the efficiency gain is directly asserted, \texttt{No} for \(q_1\) unless a financial gain is explicitly mentioned.

  \item \textbf{Multi-signal sentence (explicit-only: multiple \texttt{Yes}).}
  ``The new module runs 10$\times$ faster and offers an intuitive interface.''
  \emph{Explicit-only:} \texttt{Yes} for \(q_3\) (Organizational Impact / Performance) and \texttt{Yes} for \(q_4\) or \(q_6\) (User Well-Being / Perceived Quality), since both performance and experience are explicitly present. Under a single-label design, this creates a principled overlap case rather than annotator error.

  \item \textbf{Compliance vs.\ ``mandatory'' (explicit-only: distinguish \(q_7\) and \(q_9\)).}
  ``This procedure is required to comply with GDPR.''
  \emph{Explicit-only:} \texttt{Yes} for \(q_7\) (Regulatory Compliance). Mark \(q_9\) (Mandatory Requirement) only if the sentence explicitly frames necessity as avoiding danger/sanction or guaranteeing minimum protection (e.g., ``to avoid sanctions'' / ``to ensure minimum safety'').

  \item \textbf{Mandatory for safety (explicit-only: \(q_9=\) \texttt{Yes}).}
  ``This control is mandatory to avoid safety incidents and ensure minimum protection for users.''
  \emph{Explicit-only:} \texttt{Yes} for \(q_9\) (Mandatory Requirement) and often also \(q_8\) (Risk Prevention / Security), since danger/protection is explicitly stated.

\end{itemize}
\begin{table}[t]
\centering
\small
\setlength{\tabcolsep}{4pt}
\renewcommand{\arraystretch}{1.3}

\begin{tabular}{p{0.7cm} p{3.2cm} p{3.2cm} p{7.5cm}}
\hline
\textbf{ID} & \textbf{Category} & \textbf{Criterion name} & \textbf{Criterion (EN)} \\
\hline

$q_1$ & Performance \& Efficiency & Cost Reduction &
Does the sentence mention financial gain, cost reduction, or measurable return on investment? \\

$q_2$ & Performance \& Efficiency & Operational Efficiency &
Does the sentence highlight a clear functional improvement (time, workload, automation, \ldots) framed as an operational advantage? \\

$q_3$ & Performance \& Efficiency & Organizational Impact &
Does the sentence frame the effect as an impact on an organization's results, resources, or performance? \\

$q_4$ & User Experience \& Brand Value & User Well-Being &
Does the sentence emphasize improved well-being, comfort, quality of life, or work environment for a user? \\

$q_5$ & User Experience \& Brand Value & Reputation \& Recognition &
Does the sentence highlight a label, recognition, attractiveness, or a positive image of a service, place, or organization? \\

$q_6$ & User Experience \& Brand Value & Tangible/Perceived Quality &
Does the sentence suggest a visible, tangible, or positively perceived impact on the environment, usage, or user experience? \\

$q_7$ & Obligation \& Safety & Regulatory Compliance &
Does the sentence mention the need to comply with a standard, law, or regulatory requirement? \\

$q_8$ & Obligation \& Safety & Risk Prevention/Security &
Does the sentence refer to a security measure or risk prevention (physical, digital, legal, \ldots)? \\

$q_9$ & Obligation \& Safety & Mandatory Requirement &
Does the sentence frame an action as necessary to avoid danger, sanction, or to guarantee minimum protection? \\

\hline
\end{tabular}

\caption{PVE criteria by category (expert-defined). Each criterion is evaluated independently as \texttt{Yes}/\texttt{No} under an explicit-only annotation rule: implied value is marked \texttt{No}.}
\label{tab:criteria}
\end{table}

\section{Corpus Composition and Statistics}
\label{app:dataset-details}
\paragraph{Corpus diversity.}
Table~\ref{tab:client-stats} summarizes the corpus used for schema diagnosis. It spans five anonymized clients from distinct sectors (e.g., cybersecurity, mobility/environment, regulatory affairs, and infrastructure/energy) and includes heterogeneous document sources. The corpus is therefore constructed to capture variation in commercial writing style and procurement-relevant content rather than reflecting a single narrow domain.

\begin{table}[t]
\centering
\small
\setlength{\tabcolsep}{4pt}
\renewcommand{\arraystretch}{1.15}
\begin{tabular}{l l r r r}
\hline
\textbf{Client} & \textbf{Sector} & \textbf{\# Docs} & \textbf{\# Sents} & \textbf{Mean len.} \\
\hline
Client A & Mobility, Environment \& Well-being      & 6  & 577    & 19.02 \\
Client B & Cybersecurity \& Digital Safety         & 6  & 397    & 14.73 \\
Client C & Regulatory Affairs \& Legal             & 40 & 2{,}212 & 17.50 \\
Client D & Infrastructure, Construction \& Energy  & 4  & 1{,}227 & 18.35 \\
Client E & Buildings \& Environment                & 9  & 288    & 17.60 \\
\hline
\end{tabular}
\caption{Corpus composition for schema diagnosis by anonymized client. Counts are reported \emph{before} removing two sentences due to persistent formatting failures during response parsing (final analysis: $N=4{,}699$). Mean length is measured in whitespace-delimited tokens per sentence after sentence segmentation.}
\label{tab:client-stats}
\end{table}

\section{Prompt Design and Model Interaction}
\label{app:prompts}

\paragraph{Original prompt (English translation).}
All model queries were issued in French to match the corpus language and practitioner setting.
For readability, we provide an English translation of the shared prompt below.
The template is identical across calls: we instantiate the sentence $s\in\mathcal{S}$ and the criterion text associated with $q\in\mathcal{Q}$ (denoted $\texttt{\{sentence\}}$ and $\texttt{\{question\_text\}}$ in the template).

\begin{quote}
\small
\textit{Analysis Context -- Detection of Persuasive Value}

The sentences to be analyzed come from documents written in a B2B context
(brochures, reports, solution descriptions, etc.). These texts are part of a
persuasive communication logic, aiming to highlight the value of a solution to
a professional buyer.

Although the entire corpus pursues a persuasive objective, the analysis focuses
only on the content explicitly formulated in each sentence.
Do not take into account presumed vendor intentions or implicit inferences.

Persuasive value is an improvement, a positive effect, an advantage, or a
necessity that is clearly expressed.
If the sentence merely describes a fact, a feature, or a situation without an
explicit benefit, it does not contain persuasive value.
\medskip
\textit{Instructions:}

Read the sentence below carefully, then answer \textbf{**Oui**} or
\textbf{**Non**} to the following question.

Sentence:\\
``\{sentence\}''

Question:\\
\{question\_text\}

Respond only with \textbf{Oui} or \textbf{Non}.
\end{quote}

\paragraph{Criterion-level querying.}
Each of the nine PVE criteria ($q_1$--$q_9$, Appendix~\ref{app:criteria}) is queried independently for each sentence.
That is, for each pair $(s,q)$ with $s\in\mathcal{S}$ and $q\in\mathcal{Q}$, we issue a separate prompt instantiation, producing one binary response per (sentence, model, criterion) triple.
Treating models as annotators, this yields a response tensor (prior to cleaning) with shape $|\mathcal{S}|\times A\times Q$, where $A$ is the number of models in the panel.

\paragraph{Model panels used in the paper and appendix.}
We initially queried a panel of $A=6$ models on $|\mathcal{S}|=4{,}701$ sentences to characterize raw response formats and cleaning behavior (Appendix~\ref{app:response-cleaning}).
All main-paper diagnostics are computed on a core panel of $A=5$ models (Section~\ref{sec:experimental-setup}); the sixth model is used only for appendix-level quality and normalization analysis.

\paragraph{Interaction and decoding settings.}
All models are queried independently using the same template and each sentence is annotated in isolation (no cross-sentence context).
We use deterministic decoding (temperature $=0$) and a short output budget (max tokens $=3$) to encourage categorical responses and reduce explanatory completions.

\section{Response Collection, Cleaning, and Quality Metrics}
\label{app:response-cleaning}
    
\paragraph{Data collection.}
We queried an initial panel of $A=6$ large language models on a corpus $\mathcal{S}$ of $|\mathcal{S}|=4{,}701$ sentences using the nine PVE criteria ($q_1$--$q_9$, Appendix~\ref{app:criteria}), yielding $|\mathcal{S}|\times A\times Q = 253{,}854$ sentence--model--criterion responses (``tuples'').
All main-paper diagnostics are computed on a core panel of $A=5$ models (Section~\ref{sec:experimental-setup}).
We retain the full 6-model grid in this appendix to document response formats, cleaning, and quality statistics.

\paragraph{Normalization of raw responses.}
Models were instructed to return binary decisions in French (\textit{Oui}/\textit{Non}). In practice, surface forms varied (e.g., punctuation, markdown emphasis), and some models occasionally produced truncated positives (e.g., ``O'' as a shortened form of ``Oui''). We applied a deterministic normalization to map these variants into a boolean response tensor $\mathbf{Y}\in\{0,1\}$.
Positive variants (e.g., \textit{Oui}, \textit{Oui.}, \textit{O}, \texttt{**Oui**}, \texttt{**Oui}) were mapped to 1, and negative variants (e.g., \textit{Non}, \textit{Non.}, \texttt{**Non**}) were mapped to 0.
Table~\ref{tab:raw-types-by-model} reports the distribution of observed raw forms by model prior to normalization.
        
\begin{table}[th]
\centering
\resizebox{\textwidth}{!}{
\begin{tabular}{l r r r r r r r r r}
\hline
\textbf{Model} & \textit{Non} & \textit{Oui} & \textit{O} & \textit{Non.} & \textit{Oui.} & \texttt{**Oui**} & \texttt{**Non**} & \texttt{**Oui} & \textbf{Mal.} \\
\hline
Gemini 2.0 Flash & 36{,}370 & 5{,}609 & 0 & 0 & 0 & 277 & 53 & 0 & 0 \\
GPT-4.1-mini & 38{,}784 & 3{,}524 & 0 & 0 & 0 & 0 & 0 & 0 & 0 \\
GPT-4.1 & 38{,}988 & 3{,}270 & 0 & 50 & 1 & 0 & 0 & 0 & 0 \\
Llama-3.3-70B & 39{,}426 & 0 & 2{,}883 & 0 & 0 & 0 & 0 & 0 & 0 \\
Mistral-Large-2411 & 36{,}467 & 3{,}340 & 0 & 2{,}475 & 23 & 0 & 0 & 1 & 3 \\
Qwen-2.5-72B & 40{,}327 & 0 & 1{,}982 & 0 & 0 & 0 & 0 & 0 & 0 \\
\hline
\textbf{Total (extracted)} & 230{,}362 & 15{,}743 & 4{,}865 & 2{,}525 & 24 & 277 & 53 & 1 & 3 \\
\hline
\end{tabular}
}
\caption{Raw response types by model (non-missing tuples, before normalization). ``Mal.'' denotes malformed outputs that could not be normalized to a binary decision.}
\label{tab:raw-types-by-model}
\end{table}
    
\paragraph{Invalid tuples.}
Across the full grid, we observed four invalid tuples (0.0016\%): one missing response from \texttt{GPT-4.1-mini} and three malformed responses from \texttt{Mistral-Large-2411} that could not be mapped to a binary value (Table~\ref{tab:quality-metrics}).
        
\begin{table}[th]
\centering
\begin{tabular}{l r r r r}
\hline
\textbf{Model} & \textbf{Total} & \textbf{Missing} & \textbf{Malformed} & \textbf{Valid} \\
\hline
Gemini 2.0 Flash & 42{,}309 & 0 & 0 & 42{,}309 \\
GPT-4.1-mini & 42{,}309 & 1 & 0 & 42{,}308 \\
GPT-4.1 & 42{,}309 & 0 & 0 & 42{,}309 \\
Llama-3.3-70B & 42{,}309 & 0 & 0 & 42{,}309 \\
Mistral-Large-2411 & 42{,}309 & 0 & 3 & 42{,}306 \\
Qwen-2.5-72B & 42{,}309 & 0 & 0 & 42{,}309 \\
\hline
\textbf{TOTAL} & 253{,}854 & 1 & 3 & 253{,}850 \\
\hline
\end{tabular}
\caption{Per-model tuple quality (before filtering).}
\label{tab:quality-metrics}
\end{table}
    
\paragraph{Sentence-level filtering for a fully observed tensor.}
For downstream analyses that assume a fully observed tensor, we excluded any sentence $s\in\mathcal{S}$ for which at least one model--criterion response was missing or malformed.
This removed two sentences (IDs 2066 and 3973), yielding a filtered tensor with $|\mathcal{S}|=4{,}699$ sentences, $A=6$ models, and $Q=9$ criteria ($|\mathcal{S}|\times A\times Q = 253{,}746$ valid tuples; 100\% coverage over the filtered grid).

\begin{table}[t]
\centering
\begin{tabular}{l r}
\hline
\textbf{Metric} & \textbf{Value} \\
\hline
Sentences ($|\mathcal{S}|$) & 4{,}699 \\
Models ($A$) & 6 \\
Criteria ($Q$) & 9 \\
Total tuples ($|\mathcal{S}|\times A \times Q$) & 253{,}746 \\
\hline
Positive (1 / \textit{Oui}) & 20{,}909 (8.2\%) \\
Negative (0 / \textit{Non}) & 232{,}837 (91.8\%) \\
\hline
\end{tabular}
\caption{Filtered tensor summary ($A=6$).}
\label{tab:final-tensor-distribution}
\end{table}

\paragraph{Dropped sentences.}
Table~\ref{tab:dropped-sentences-text} reports the exact text of the two removed sentences (French original and English translation). Sentence 2066 (``EXEMPLE DE TRAME RE\c{C}UE'') appears to have triggered malformed output in \texttt{Mistral-Large-2411}. Sentence 3973, a long administrative title, resulted in a single missing response from \texttt{GPT-4.1-mini}.
        
\begin{table}[t]
\centering
\begin{tabular}{p{1.2cm} p{6.2cm} p{6.2cm}}
\hline
\textbf{ID} & \textbf{French (original)} & \textbf{English (translation)} \\
\hline
2066 & EXEMPLE DE TRAME RECUE & EXAMPLE OF TEMPLATE RECEIVED \\
\hline
3973 & Diagnostic agricole dans le cadre du Plan local d'urbanisme intercommunal de la Communaut\'e de communes de la r\'egion de Suippes &
Agricultural diagnostic within the framework of the inter-municipal local urban planning scheme of the Suippes Region Community of Communes \\
\hline
\end{tabular}
\caption{Text of dropped sentences (French original and English translation).}
\label{tab:dropped-sentences-text}
\end{table}

\section{Model Panel Details}
\label{app:model-panel}

\paragraph{Models used.}
All main-paper diagnostics are computed on a core panel of $A=5$ models (treated as annotators in $\mathbf{Y}$):

\begin{itemize}

    \item \texttt{openai/gpt-4.1-mini} (OpenAI)
    \item \texttt{openai/gpt-4.1} (OpenAI)
    \item \texttt{meta-llama/llama-3.3-70b-instruct} (Meta, via OpenRouter)
    \item \texttt{mistralai/mistral-large-2411} (Mistral, via OpenRouter)
    \item \texttt{qwen/qwen-2.5-72b-instruct} (Qwen, via OpenRouter)
\end{itemize}

We additionally report response format and normalization statistics for the
\texttt{gemini-2.0-flash} model (via OpenRouter), this model is not used in the main diagnostic results.

\paragraph{Panel rationale.}
We deliberately select models from different families/providers to reduce dependence on any single training lineage or instruction style, and to probe whether stability/overlap patterns persist across heterogeneous LLM annotators.

\paragraph{Shared decoding settings.}
All models are queried independently with deterministic decoding (temperature $=0$) and a short output budget (max tokens $=3$). Each criterion is prompted in a separate call, and each sentence is annotated in isolation (no cross-sentence context).

\paragraph{On the use of LLMs as annotators.}
Using LLMs as annotators is an acknowledged limitation and an open research 
question. LLM judgments may reflect shared training priors or similar 
instruction-following behavior, which can shape observed instability and 
overlap patterns. However, this concern applies equally to human panels: 
annotators sharing the same professional background, institutional context, 
or domain expertise may similarly align in their operationalizations, 
introducing correlated variation that is not unique to LLMs. Panel homogeneity 
is a general issue in multi-annotator settings, and the framework does not 
assume annotator independence; rather, it is designed to quantify how 
disagreement structures vary as panel composition changes. In this sense, 
sensitivity to annotator diversity, whether human or LLM, is not an 
uncontrolled confound, but an observable and empirically measurable property 
of the diagnostic. The human validation subset, which operates at the category 
level rather than the criterion level, provides a complementary external anchor: 
alignment between diagnostic signals and expert category-level disagreement 
confirms that instability and overlap reflect schema structure rather than 
LLM-specific behavior.

\section{Human Annotation Setup and Reliability}
\label{app:annotation-details}

\paragraph{Expert panel and task.}
Five domain experts with procurement experience annotated a validation subset using the intended task framing: each sentence receives exactly one label from $\mathcal{C}=\{c_0,c_1,c_2,c_3\}$ (Non-Persuasive, Performance \& Efficiency, User Experience \& Brand Value, Obligation \& Safety). The category definitions were the outcome of prior calibration discussions with these experts during schema development: they jointly agreed on the category meanings and the underlying criterion decomposition used in the diagnostic (Appendix~\ref{app:criteria}).

\paragraph{Annotation instructions.}
Experts were instructed to label sentences based on what is explicitly stated in the sentence (explicit-only principle, Appendix~\ref{app:criteria}), without relying on unstated assumptions or inferred vendor intent. For the human validation, experts did \emph{not} answer the nine criteria directly, instead, they assigned a single category label per sentence, consistent with the intended downstream single-label task design. This choice reduces annotation burden and reflects how practitioners apply the schema holistically in procurement workflows, while still allowing us to test whether criterion-level instability/overlap predicts where experts disagree.

\paragraph{Sampling and accountability.}
We construct a validation set of 500 annotation \emph{assignments} designed to reflect corpus diversity rather than maximize i.i.d.\ representativeness. The set covers multiple clients/sectors and includes a mix of diagnostic profiles (e.g., clear non-persuasive, clear persuasive, and multi-dimensional cases where multiple criteria/categories are engaged by the diagnostic). This stratified construction ensures that human validation includes both easy reference cases and challenging boundary cases that drive the schema-level conclusions.

\paragraph{Unique sentences and repeated items.}
The validation set contains 389 unique sentences plus 111 repeated sentences (i.e., the same sentence shown twice to the same expert) used to estimate within-expert consistency under the single-label task. We report test--retest consistency as the fraction of repeated items for which an expert assigns the same category both times (equivalently, $1-\textit{flip rate}$).

\paragraph{Within-expert (test--retest) reliability.}
Across experts, each annotated 62 repeated items. Test--retest consistency varies across experts (approximately 0.71--0.90 in our runs), with flips concentrating in diagnostically complex cases (e.g., multi-dimensional overlap) rather than in clear non-persuasive items. This pattern is consistent with the main claim that disagreement is structured around specific boundaries and overlap regimes, not diffuse annotation noise.

\begin{figure}[t]
  \centering
  \includegraphics[width=0.45\columnwidth]{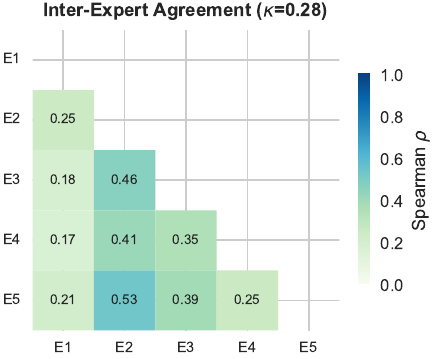}
  \caption{Pairwise inter-expert agreement (binary agreement on whether two experts assign the same category), with overall Fleiss' $\kappa$ reported for the expert panel.}
  \label{fig:human-iaa-heatmap}
\end{figure}

\paragraph{Between-expert agreement.}
We quantify inter-expert agreement using Fleiss' $\kappa$ over the categorical labels. Agreement is moderate-to-low ($\kappa \approx 0.28$ in our setting), reflecting the inherent difficulty of forced single-label assignment for sentences that can support multiple procurement-relevant rationales. To localize this disagreement, we also visualize pairwise agreement between experts (Figure~\ref{fig:human-iaa-heatmap}), which shows that agreement varies substantially by expert pair, again consistent with the task admitting multiple defensible readings rather than a single latent ground truth.

\paragraph{Interpretation.}
These reliability statistics are not used to adjudicate a ``correct'' label. Instead, they serve as an external anchor for the diagnostic: items flagged as multi-category by the criterion-level audit are also the items where experts are most likely to diverge under forced single-label assignment, supporting the interpretation that observed overlap reflects a schema/taxonomy bottleneck rather than random annotation error.

\section{Robustness to Engagement Threshold}
\label{app:robustness-rules}

\paragraph{Motivation.}
All diagnostics in Section~\ref{sec:method} are computed \emph{conditional on engagement}, using focus sets
$\Omega_{q,t}=\{\, s\in\mathcal{S} \mid v_{sq}\ge t \,\}$ (Eq.~\ref{eq:omega}), where
$v_{sq}=\sum_{a=1}^{A} y_{saq}$ is the number of positive votes (out of $A$) for criterion $q$ on unit $s$ (Eq.~\ref{eq:votes}).
Because criteria are sparse (many sentences are purely descriptive), summary statistics computed over the full corpus would be dominated by shared \textit{non-activation} rather than meaningful agreement \emph{when a criterion is actually applicable}.
A natural concern is that the stability patterns reported in the main analysis could depend on the engagement threshold $t$
(e.g., because $t=1$ admits singleton endorsements).

To assess robustness, we recompute all quantities under multiple thresholds:
\[
t=1
\qquad
t=2
\qquad
t=\left\lceil\frac{A}{2}\right\rceil,
\]
which correspond (for $A=5$) to $t\in\{1,2,3\}$.
Each threshold induces a criterion-specific focus set $\Omega_{q,t}$ and activation rate
$\mathrm{Act}_{t}(q)=|\Omega_{q,t}|/|\mathcal{S}|$ (Eq.~\ref{eq:act}).

\paragraph{Agreement structure under different thresholds.}
For each criterion $q$ and threshold $t$, we recompute the conditional vote-count distribution
$\pi_{q}(k\mid t)$ (Eq.~\ref{eq:pi}) over $k\in\{t,\dots,A\}$.
To summarize boundary pressure with a single scalar, we define an \textbf{ambiguity rate} as the mass of near-ties among engaged units.
For $A=5$, near-ties correspond to vote counts $\{2,3\}$, so we define:
\[
\mathrm{Amb}_{t}(q)=\pi_{q}(2\mid t)+\pi_{q}(3\mid t).
\]
Note that $\mathrm{Amb}_{t}$ is always computed \emph{within} the threshold-specific focus set $\Omega_{q,t}$; increasing $t$ changes which units enter the conditioning set, not the definition of ``near-tie'' itself.

\begin{figure}[t]
\centering
\includegraphics[width=0.95\columnwidth]{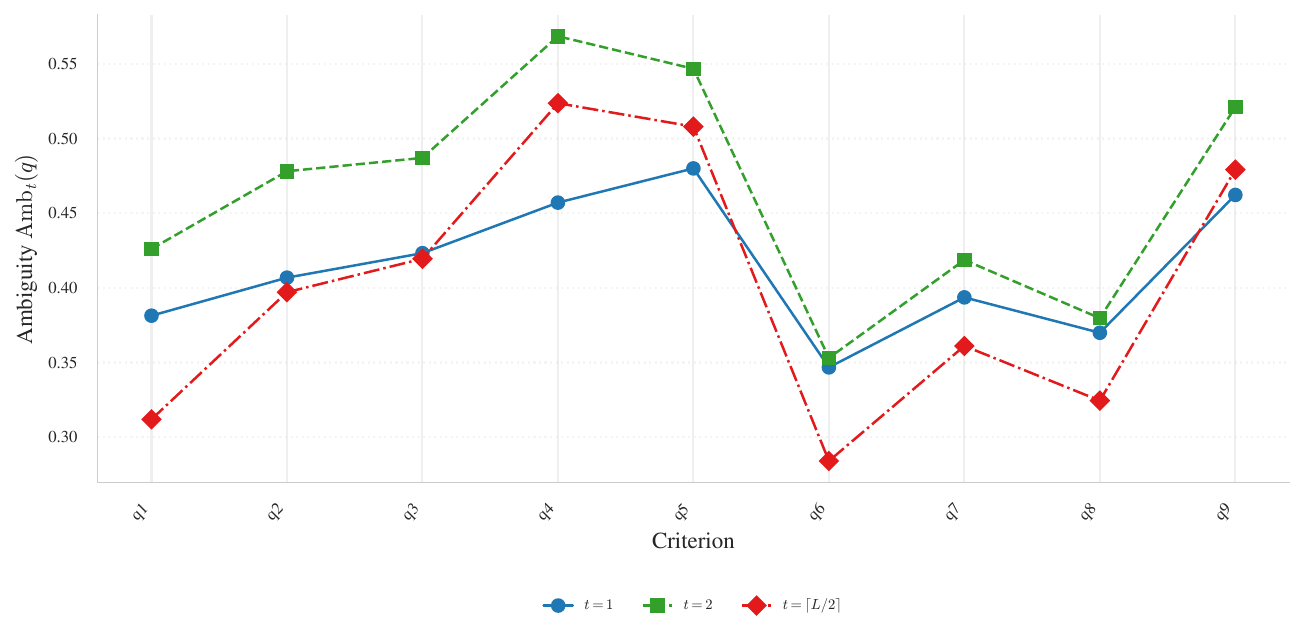}
\caption{Sensitivity of the ambiguity profile to threshold choice (computed conditional on the threshold-specific focus set $\Omega_{q,t}$).
The relative ordering of criteria is preserved across $t=1$, $t=2$, and $t=\lceil A/2\rceil$, indicating threshold-robust hotspot identification.}
\label{fig:rule-amb}
\end{figure}

\begin{figure}[t]
\centering
\includegraphics[width=0.95\columnwidth]{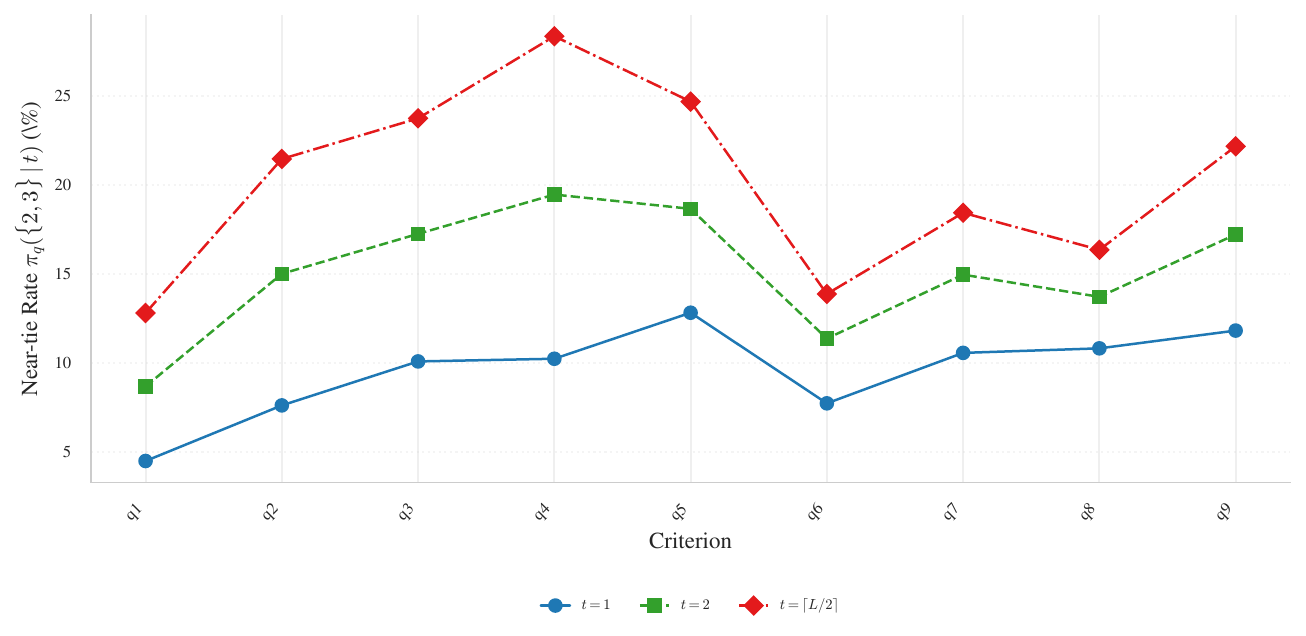}
\caption{Sensitivity of near-tie rates to threshold choice (conditional on $\Omega_{q,t}$). For $A=5$, the near-tie rate equals $\mathrm{Amb}_t(q)=\pi_{q}(2\mid t)+\pi_{q}(3\mid t)$.}
\label{fig:rule-nt}
\end{figure}

\paragraph{Why ambiguity can increase under stricter thresholds.}
$\mathrm{Amb}_{t}(q)$ can increase from $t=1$ to stricter regimes due to a \textbf{compositional shift} induced by conditioning.
Under $t=1$, $\Omega_{q,1}$ includes many weak activations (e.g., singleton endorsements), which can dilute the share of borderline splits.
Stricter thresholds filter to units with stronger multi-annotator engagement; in subjective settings, these are often precisely the most contentious units and therefore more likely to concentrate around 2--3 / 3--2 splits.
\paragraph{Robustness summary (ambiguity + activation).}
Across thresholds, stricter engagement reduces coverage (lower $\mathrm{Act}_{t}(q)$), but the qualitative picture of which criteria produce borderline splits is stable:
the same criteria remain disproportionately ambiguous, and hotspot identification is not an artifact of a particular threshold choice
(Figure~\ref{fig:rule-amb}, Figure~\ref{fig:rule-nt}, Table~\ref{tab:rule-robustness-full}).

\begin{table}[t]
\centering
\small
\begin{tabular}{l ccc ccc cc}
\hline
& \multicolumn{3}{c}{\textbf{Ambiguity ($\mathrm{Amb}_t$)}} & \multicolumn{3}{c}{\textbf{Activation (Act$_t$, \%)}} & & \\
\cline{2-4}\cline{5-7}
\textbf{Criterion} & $t=1$ & $t=2$ & $t=\lceil A/2\rceil$ & $t=1$ & $t=2$ & $t=\lceil A/2\rceil$ & \textbf{Rank Range} & \textbf{Top-3} \\
\hline
$q_5$ & 0.480 & 0.547 & 0.508 & 20.56 & 14.13 & 10.68 & 1--2 & 3/3 \\
$q_9$ & 0.462 & 0.521 & 0.479 & 12.77 &  8.77 &  6.81 & 2--3 & 3/3 \\
$q_4$ & 0.457 & 0.568 & 0.524 & 17.02 &  8.96 &  6.15 & 1--3 & 3/3 \\
\hline
$q_3$ & 0.423 & 0.487 & 0.419 & 19.17 & 11.22 &  8.15 & 4--4 & 0/3 \\
$q_2$ & 0.407 & 0.478 & 0.397 & 17.85 &  9.07 &  6.34 & 5--5 & 0/3 \\
$q_7$ & 0.394 & 0.419 & 0.361 &  6.24 &  4.41 &  3.58 & 6--7 & 0/3 \\
$q_1$ & 0.381 & 0.426 & 0.312 &  4.72 &  2.45 &  1.66 & 6--8 & 0/3 \\
$q_8$ & 0.370 & 0.380 & 0.324 & 14.54 & 11.47 &  9.62 & 7--8 & 0/3 \\
$q_6$ & 0.347 & 0.353 & 0.284 & 34.35 & 23.37 & 19.15 & 9--9 & 0/3 \\
\hline
\end{tabular}
\caption{Robustness of ambiguity hotspots to threshold choice.
Ambiguity is defined as near-tie mass $\mathrm{Amb}_{t}(q)=\pi_q(2\mid t)+\pi_q(3\mid t)$ for $A=5$.
While coverage decreases under stricter thresholds (lower Act$_t$), the identity of the most ambiguous criteria is stable:
$q_5$, $q_9$, and $q_4$ appear in the top-3 ambiguity ranking under all three regimes (Top-3 frequency $=3/3$).}
\label{tab:rule-robustness-full}
\end{table}

\begin{figure}[t]
\centering
\includegraphics[width=0.95\columnwidth]{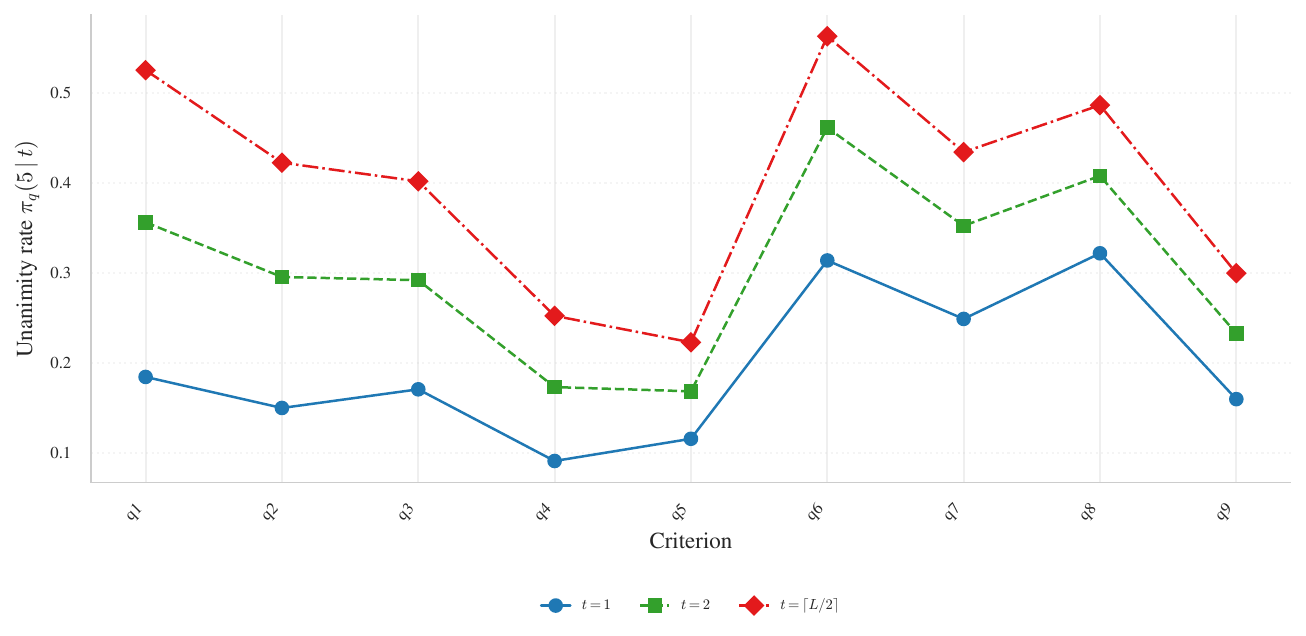}
\caption{Sensitivity of unanimity rates to threshold choice (computed conditional on the threshold-specific focus set $\Omega_{q,t}$).
The plot reports $\mathrm{UY}_t(q)=\pi_q(A\mid t)$ for each criterion under $t=1$, $t=2$, and $t=\lceil A/2\rceil$ ($A=5$).
Despite reduced coverage under stricter thresholds, the relative ordering of criteria is preserved:
criteria that are crisp at $t=1$ (notably $q_6$) remain highly unanimous, while ambiguity hotspots remain low-unanimity.}

\label{fig:rule-uy}
\end{figure}

\paragraph{Unanimity robustness.}
The main paper also interprets stability through the unanimity rate, which captures how often a criterion yields full agreement \emph{when engaged}.
We therefore recompute the unanimity rate
$\mathrm{UY}_t(q)=\pi_q(A\mid t)$ across thresholds.
Figure~\ref{fig:rule-uy} shows the same qualitative pattern as the ambiguity analysis:
criteria that are comparatively crisp at $t=1$ (notably $q_6$) remain consistently more unanimous across $t\in\{1,2,\lceil A/2\rceil\}$,
while ambiguity hotspots (e.g., $q_4$, $q_5$, $q_9$) remain low-unanimity.
Together, these checks indicate that the main stability findings are not artifacts of the permissive $t=1$ regime.

\section{Robustness to Model Selection}
\label{app:robustness-panel}

\paragraph{Motivation.}
A natural concern is whether the criterion-specific instability patterns reported in the main paper
depend on the particular models chosen for the panel (e.g., one model being systematically stricter or looser),
rather than reflecting properties of the criterion definitions interacting with the corpus.
We therefore test whether the criterion-level signatures persist under controlled perturbations of panel composition.

\paragraph{Fixed-size leave-one-model-out design.}
To preserve the voting geometry, we perform a \textbf{leave-one-model-out (LOO)} analysis at a \emph{fixed} panel size $A=5$.
We temporarily expand the candidate pool to six models by adding \texttt{gemini-2.0-flash-001} solely for this check, yielding $A^{\star}=6$ candidate models and thus six distinct 5-model panels (5-of-6), each obtained by removing one model.
Keeping $A$ fixed avoids confounding effects from changing the probability of ties or near-ties and ensures that all agreement quantities (e.g., near-tie mass at $\{2,3\}$ when $A=5$) remain directly comparable across panels.

Crucially, the ablations also probe \emph{model-specific bias}: different models can be systematically more permissive or conservative, and they differ in training lineage, prompting style, and parameter scale.
By rotating which model is excluded while holding the aggregation and conditioning rules constant, we test whether the criterion-level patterns persist under controlled perturbations of panel composition rather than being driven by any single model’s idiosyncratic thresholding.
We include Gemini as an additional, lightweight but frontier-level model from a different family to broaden the pool and reduce the risk that the robustness check merely re-tests highly similar model behaviors.

\paragraph{Recomputed quantities (same notation as the main paper).}
Let $y_{saq}\in\{0,1\}$ denote the vote of model $a$ on unit $s\in\mathcal{S}$ for criterion $q\in\mathcal{Q}$, and let
$v_{sq}=\sum_{a=1}^{A} y_{saq}$ be the number of positive votes under a given 5-model panel.
For each ablated panel, we recompute the same conditional agreement statistics as in the main analysis using focus sets
\[
\Omega_{q,t}=\{\, s\in\mathcal{S} \mid v_{sq}\ge t \,\},
\]
and we report results under the permissive engagement rule $t=1$ (triggered-at-least-once).
Within each $\Omega_{q,1}$, we recompute the conditional vote-count distribution
$\pi_q(k\mid 1)$ over $k\in\{1,\dots,A\}$ and its derived summaries (NT, AS, UY).

\begin{figure}[th]
\centering
\includegraphics[width=0.95\columnwidth]{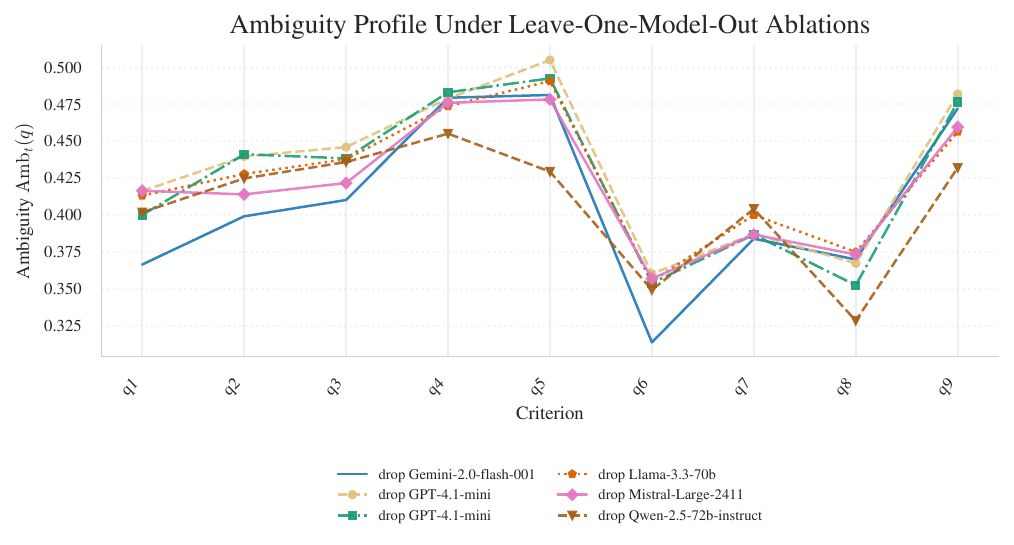}
\caption{Leave-one-model-out robustness of near-tie rates at $t=1$ ($A=5$).
Each curve recomputes $\mathrm{NT}(q)=\pi_q(2\mid 1)+\pi_q(3\mid 1)$ after dropping one model from the 6-model candidate pool ($A^{\star}=6$).
The qualitative criterion ordering is preserved across ablations, indicating that boundary-mass patterns are not driven by a single model.}
\label{fig:loo-nt}
\end{figure}

\paragraph{Results: near-tie profiles are stable across ablations.}
Because the main paper interprets instability through mass near the decision boundary, we first examine the near-tie rate NT derived from $\pi_q(\cdot\mid 1)$.
For $A=5$, near-ties correspond to $k\in\{2,3\}$, so
$\mathrm{NT}(q)=\pi_q(2\mid 1)+\pi_q(3\mid 1)$.
Figure~\ref{fig:loo-nt} overlays $\mathrm{NT}(q)$ for all six 5-model panels (each curve drops one model from the 6-model pool).
The profiles are qualitatively similar across ablations, indicating that the criteria exhibiting high boundary mass under the main panel remain the same under panel perturbations.

\paragraph{Audit table: rank stability under model ablations.}
To make robustness easy to verify, Table~\ref{tab:loo-nt-summary} reports, for each criterion $q$,
(i) the near-tie rate under the \emph{main} panel (the original five-model panel),
(ii) the min--max range across LOO panels, and
(iii) how often the criterion appears among the three largest NT values across the six ablations.

\begin{table}[th]
\centering
\small
\begin{tabular}{@{}lcccc@{}}
\hline
\textbf{Criterion} & \textbf{NT\textsubscript{main}} & \textbf{Range} & \textbf{$\Delta$NT (pp)} & \textbf{Top-3 freq.} \\
\hline
$q_5$ & 30.7\% & 30.7--38.4\% & 7.7 & 6/6 \\
$q_9$ & 29.5\% & 29.5--38.2\% & 8.7 & 6/6 \\
$q_4$ & 27.3\% & 27.3--35.8\% & 8.5 & 4/6 \\
$q_7$ & 28.5\% & 23.7--28.5\% & 4.8 & 1/6 \\
$q_3$ & 28.4\% & 24.0--29.8\% & 5.8 & 1/6 \\
\hline
\end{tabular}
\caption{Near-tie robustness across 5-of-6 LOO panels at $t=1$ ($A=5$).
NT$_{\mathrm{main}}$ is computed on the main panel (the original five-model panel).
``Top-3 freq.'' counts how often each criterion appears among the three largest NT values across the six ablations.}
\label{tab:loo-nt-summary}
\end{table}

\begin{figure}[t]
\centering
\includegraphics[width=0.95\columnwidth]{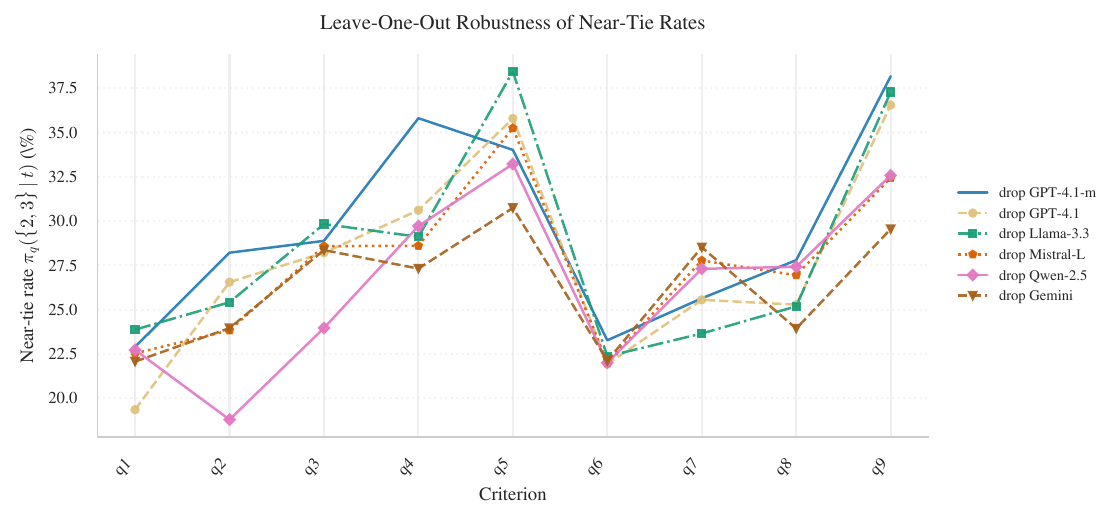}
\caption{Leave-one-model-out robustness of unanimity rates at $t=1$ ($A=5$).
Each curve recomputes $\mathrm{UY}(q)=\pi_q(A\mid 1)$ after dropping one model from the 6-model candidate pool ($A^{\star}=6$).
The preserved criterion ordering indicates that ``crisp'' versus ``ambiguous'' behavior is not panel-specific.}
\label{fig:loo-uy}
\end{figure}

\paragraph{Complementary view: unanimity ordering is stable.}
We also verify that the criteria identified as comparatively crisp in the main analysis remain so under panel perturbations.
Specifically, we recompute the unanimity rate $\mathrm{UY}(q)=\pi_q(A\mid 1)$ for each LOO panel.
Figure~\ref{fig:loo-uy} shows that the relative ordering is preserved: criteria with high unanimity under the main panel remain high-unanimity across ablations, while low-unanimity criteria remain low.
This supports the interpretation that panel perturbations do not qualitatively change which criteria behave as stable measurement instruments.

\paragraph{Implications for schema diagnosis.}
Across all 5-of-6 ablations, both the near-tie profile (NT) and the unanimity profile (UY) remain qualitatively stable.
When multiple model backends yield similar criterion-specific boundary patterns under fixed $A$ and identical conditioning,
the most plausible explanation is that these patterns reflect properties of the criterion definitions and their operational boundaries,
rather than idiosyncrasies of a single model in the panel.
This robustness strengthens the main paper’s claim that subjectivity is diagnosable at the criterion level.
We also verified that the activation profile $\mathrm{Act}_{1}(q)=|\Omega_{q,1}|/|\mathcal{S}|$ remains qualitatively similar across ablations,
so the stability of NT and UY is not explained by drastic changes in coverage.

\section{Additional Overlap Diagnostics and Robustness}
\label{app:overlap}

\paragraph{Why we mask within-category blocks in the main figure.}
In the main paper, we visualize \emph{cross-category} conditional overlap to directly stress-test the schema’s intended separations.
Within-category co-engagement is expected because multiple criteria may support the same category, and these within-block interactions can visually dominate the matrix.
We therefore mask within-category blocks in Figure~\ref{fig:overlap_heatmap} to emphasize \emph{boundary blurring} across categories.
This appendix provides the corresponding unmasked view for auditability.

\begin{figure}[t]
  \centering
  \includegraphics[width=0.5\columnwidth]{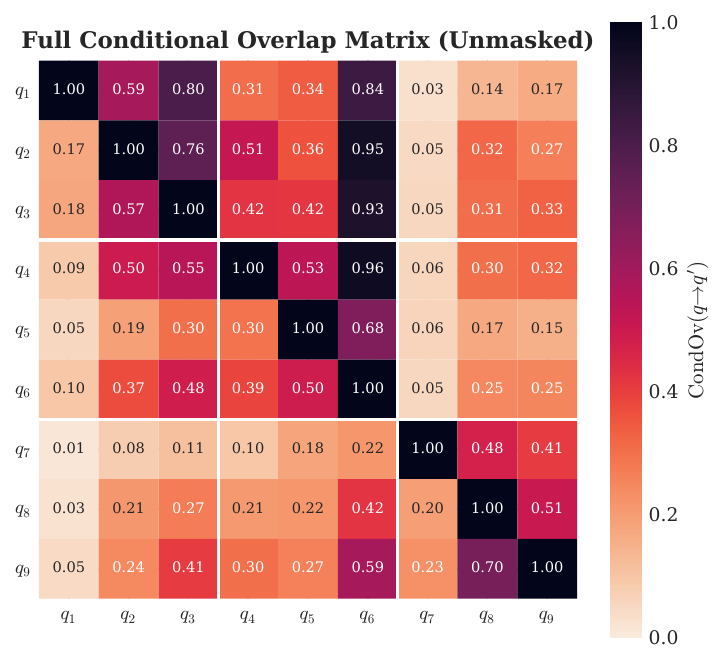}
  \caption{Full conditional overlap matrix at $t=1$ (unmasked).
  Cell $(q,q')$ reports the directed conditional overlap $\mathrm{CondOv}_{1}(q\!\rightarrow\! q')$ (Eq.~\ref{eq:condov}).}
  \label{fig:condov_full}
\end{figure}

\paragraph{Full conditional overlap structure.}
Figure~\ref{fig:condov_full} reports the complete conditional overlap matrix at $t=1$.
The matrix confirms that within-category co-engagement can be substantial, but it also reveals strong cross-category dependencies.
In particular, several criteria exhibit broad cross-category co-engagement patterns (i.e., when they are engaged, multiple criteria from other categories are frequently engaged), consistent with structured rather than diffuse overlap.

\paragraph{Sensitivity to engagement thresholding.}
A natural concern is that overlap could be inflated by permissive engagement (e.g., retaining single-annotator activations at $t=1$).
To test robustness, Table~\ref{tab:overlap_rule_sensitivity} recomputes overlap under stricter engagement thresholds
$t\in\{1,2,\lceil A/2\rceil\}$ (for $A=5$, $t\in\{1,2,3\}$).
As expected, stricter thresholds reduce coverage and lower overlap rates by filtering weaker activations.
However, cross-category co-activation among covered units remains substantial across thresholds, indicating that boundary blurring is not an artifact of $t=1$ but a stable property of the schema on units where the schema applies.

\begin{table}[t]
\centering
\small
\setlength{\tabcolsep}{7pt}
\renewcommand{\arraystretch}{1.05}
\begin{tabular}{l r r c c}
\hline
\textbf{Threshold $t$} & \textbf{Covered} & \textbf{Coverage} & $\mathbf{\mathrm{Overlap}_{\mathrm{cat}\mid \mathrm{cov},t}}$ & $\mathbf{\Pr(|\Gamma_{s,t}|\ge 2)}$ \\
\hline
$1$ & 1{,}951 & 41.5\% & 44.6\% & 26.8\% \\
$2$ & 1{,}605 & 34.2\% & 39.1\% & 19.9\% \\
$3$ & 1{,}358 & 28.9\% & 36.4\% & 16.3\% \\
\hline
\end{tabular}
\caption{Engagement robustness for overlap diagnostics.
\emph{Covered} counts units with at least one active non-target category ($m_{s,t}\ge 1$).
$\mathrm{Overlap}_{\mathrm{cat}\mid \mathrm{cov},t}$ is the fraction of covered units that co-activate at least two categories ($m_{s,t}\ge 2$, Eq.~\ref{eq:overlap-cat-cov}).
$\Pr(|\Gamma_{s,t}|\ge 2)$ measures criterion-level multi-signal engagement (Eq.~\ref{eq:gamma}).}
\label{tab:overlap_rule_sensitivity}
\end{table}

\begin{figure}[t]
  \centering
  \includegraphics[width=\columnwidth]{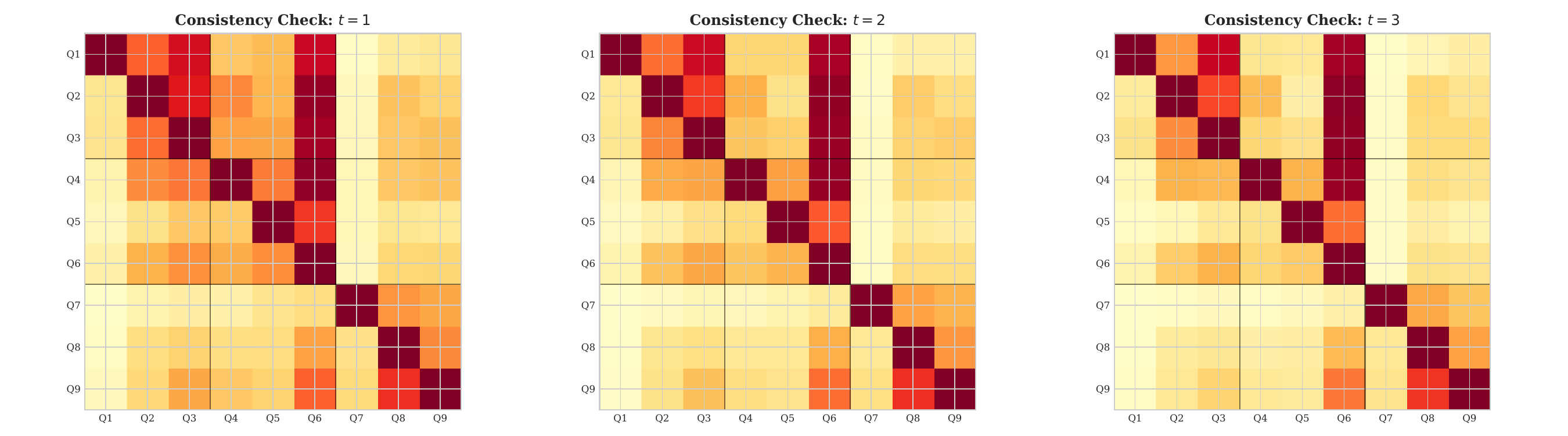}
  \caption{Consistency check for overlap structure across engagement thresholds ($t=1,2,3$).
  Each panel reports $\mathrm{CondOv}_{t}(q\!\rightarrow\! q')$ (Eq.~\ref{eq:condov}).
  Strong cross-category interactions remain visible across thresholds, while weaker links diminish under stricter engagement.}
  \label{fig:overlap_consistency}
\end{figure}

\paragraph{Consistency of overlap structure across thresholds.}
Beyond aggregate rates, we test whether the \emph{structure} of overlap is stable under stricter engagement.
Figure~\ref{fig:overlap_consistency} shows conditional overlap matrices for $t\in\{1,2,3\}$.
The qualitative pattern is consistent: the strongest cross-category interactions persist, while weaker links fade as $t$ increases.

\paragraph{Directed asymmetry: subset-like behavior persists across thresholds.}
Overlap is often markedly \emph{asymmetric}, consistent with subset-like behavior rather than symmetric entanglement (Eq.~\ref{eq:condov}).
As a representative example, Table~\ref{tab:asymmetry_q2_q6} reports the pair $(q_2,q_6)$ across thresholds.
The asymmetry is stable: $q_6$ is almost always engaged when $q_2$ is engaged, but the reverse is substantially weaker, suggesting that $q_2$ frequently appears as a more specific instance or proxy of the broader signal captured by $q_6$.

\begin{table}[t]
\centering
\small
\setlength{\tabcolsep}{10pt}
\renewcommand{\arraystretch}{1.05}
\begin{tabular}{lcc}
\hline
\textbf{Threshold $t$} & $\mathbf{\mathrm{CondOv}_{t}(q_2\!\rightarrow\! q_6)}$ & $\mathbf{\mathrm{CondOv}_{t}(q_6\!\rightarrow\! q_2)}$ \\
\hline
$1$ & 0.95 & 0.37 \\
$2$ & 0.96 & 0.32 \\
$3$ & 0.97 & 0.29 \\
\hline
\end{tabular}
\caption{Directional asymmetry example for $(q_2,q_6)$ across engagement thresholds.
Values are directed conditional overlaps $\mathrm{CondOv}_{t}(q\!\rightarrow\! q')$ (Eq.~\ref{eq:condov}).}
\label{tab:asymmetry_q2_q6}
\end{table}

\section{Model Panel Sensitivity and Annotator Bias}
\label{app:model-behavior}

\begin{figure}[ht]
\centering
\includegraphics[width=0.7\columnwidth]{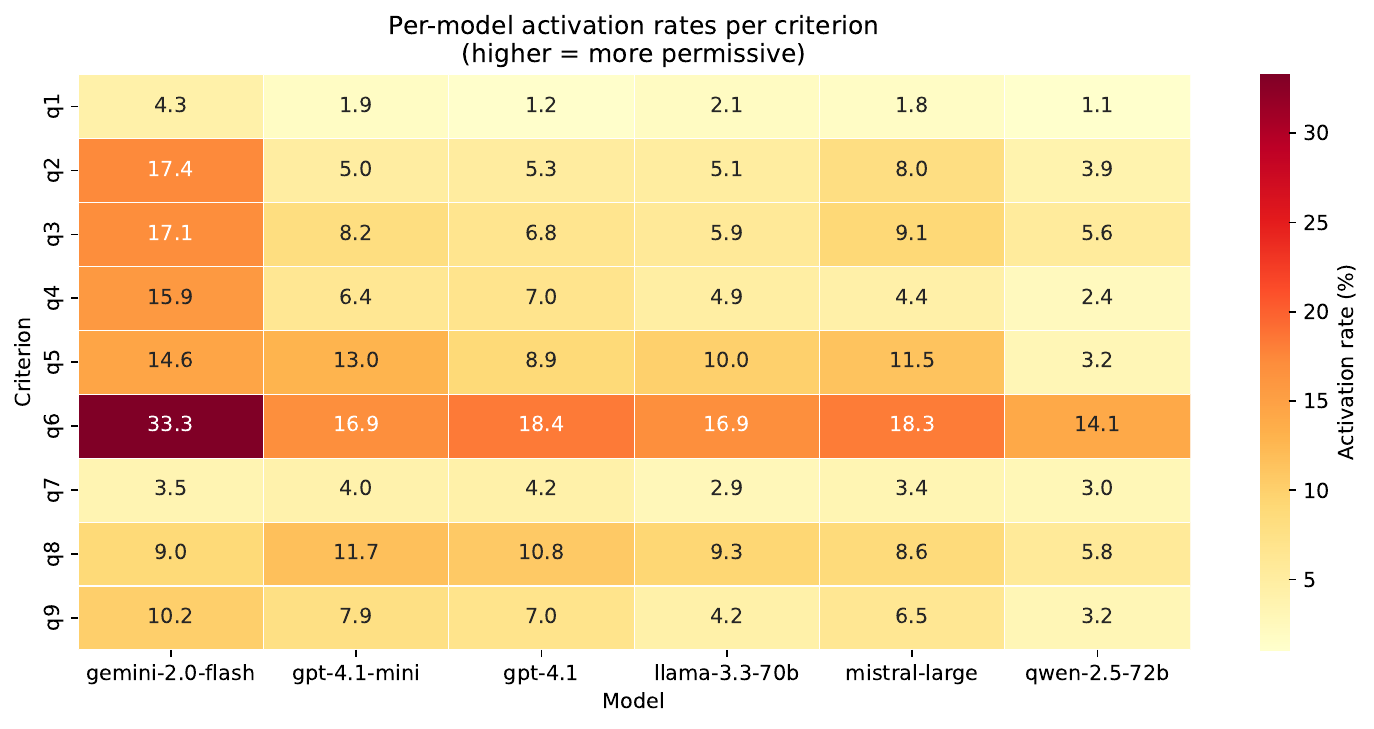}
\caption{Per-model activation rates per criterion. Higher values 
indicate a more permissive model for that criterion. Activation 
rates are corpus-level (fraction of all sentences where the model 
voted yes). Gemini shows substantially higher rates on 
experiential criteria ($q_4$--$q_6$), while Qwen is 
consistently conservative.}
\label{fig:model_bias}
\end{figure}

\paragraph{Construct-dependent model behavior.}
A key assumption behind multi-model annotation panels is that 
model diversity introduces heterogeneous operationalizations of 
the criteria, making the diagnostic more robust than any single 
annotator's judgment. The per-model activation profiles 
(Figure~\ref{fig:model_bias}) confirm that this heterogeneity 
is real but construct-dependent rather than uniform. Gemini 
activates criteria related to user experience and brand value 
($q_4$--$q_6$) at substantially higher rates than other models, 
while remaining conservative on obligation-related criteria 
($q_7$--$q_9$). Qwen consistently shows the lowest activation 
rates across all criteria. These are not random differences: 
they reflect how each model's training and instruction-following 
style interact with the semantic nature of each criterion.

This pattern is not unique to LLMs. Human annotators similarly 
exhibit construct-dependent thresholds: an annotator with a 
legal background may apply compliance criteria more strictly 
than one with a marketing background, while showing similar 
behavior on performance criteria. Model bias, like human bias, 
is not a uniform property of the annotator but a property of 
the annotator-criterion interaction. Acknowledging this 
construct-dependence is more informative than simply asserting 
that a panel is diverse.

\begin{figure}[ht]
\centering
\includegraphics[width=\columnwidth]{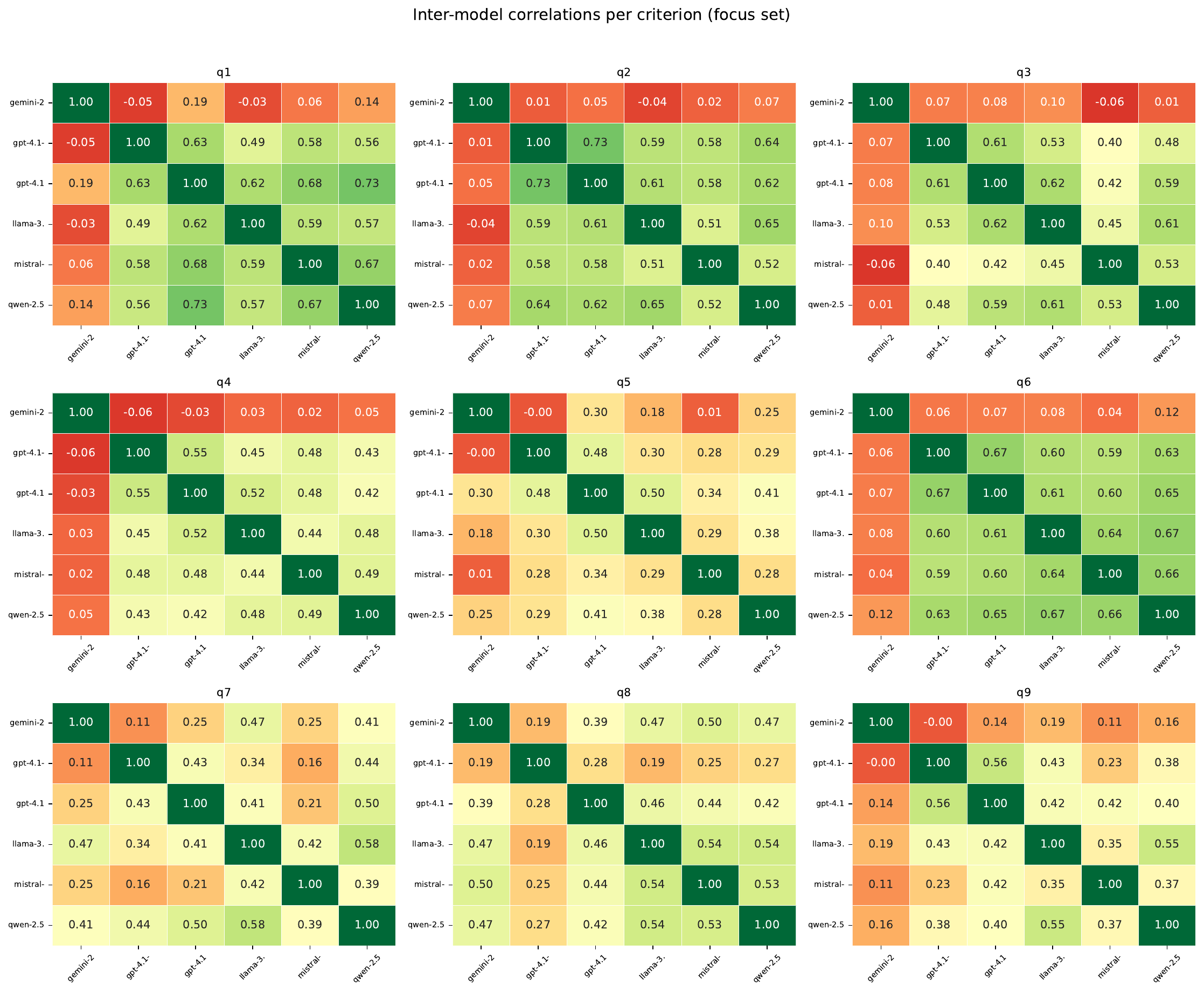}
\caption{Inter-model pairwise correlations per criterion, 
computed on the focus set $\Omega_{q,1}$. No model pair 
aligns consistently across all nine criteria. Gemini shows 
near-zero or negative correlations on unstable criteria 
($q_4$, $q_5$, $q_9$), while correlations among the four 
non-Gemini models are moderate and heterogeneous 
($\approx 0.10$--$0.73$ depending on the criterion).}
\label{fig:model_correlations}
\end{figure}

\paragraph{No model pair aligns consistently across all criteria.}
Inter-model correlation matrices per criterion 
(Figure~\ref{fig:model_correlations}) show that correlations 
range from approximately $-0.06$ to $0.73$ depending on the 
criterion, with no model pair consistently aligning across 
all nine criteria. Gemini shows near-zero or negative 
correlations with other models on unstable criteria 
($q_4$, $q_5$, $q_9$), precisely where disagreement is 
most diagnostically informative. This heterogeneity is what 
makes a diverse panel valuable: instability that persists 
despite model disagreement is more likely to reflect genuine 
criterion ambiguity than any single model's idiosyncratic 
threshold.

\paragraph{Implications and open directions.}
The observation that model behavior varies by construct opens 
a broader methodological question: rather than a fixed panel 
applied uniformly across all criteria, one could in principle 
select models per criterion based on their semantic alignment 
with the construct being measured. A model that applies 
compliance criteria conservatively and precisely might be 
more informative for $q_7$--$q_9$, while a more permissive 
model might better surface boundary cases for $q_5$. This 
criterion-adaptive panel design is an open research direction.

More broadly, the results reinforce that model selection is 
itself a design decision with epistemic consequences. Reporting 
which models were used, how their activation profiles differ, 
and where they agree or disagree should become standard 
practice in LLM-based schema auditing. The diagnostic framework 
proposed in this paper is annotator-agnostic by design, but 
the specific signals it surfaces are shaped by the panel 
composition. Making this sensitivity explicit, as we do here, 
is part of what makes the diagnostic interpretable rather 
than opaque.

\section{Schema Refinement in Practice}
\label{app:refinement}

\paragraph{From diagnostic signal to targeted revision.}
The diagnostic identified $q_5$ (Reputation \& Recognition) as the 
most structurally problematic criterion: it exhibited both high 
near-tie rates and systematic overlap with $q_6$ (Perceived Quality). 
Manual analysis of sentences activated by $q_5$ revealed the root 
cause: the criterion conflated two fundamentally different signal 
families. The first captures \emph{evidence-based credibility}, 
explicit labels, certifications, awards, and rankings, which are 
verifiable and trigger stable annotator agreement. The second captures 
\emph{claim-based reputation}, positive image, attractiveness, 
prestige, which are subjective and highly open to inference. When 
both families are grouped under a single criterion, strict annotators 
accept only verifiable proof while permissive annotators also accept 
implicit image signals, producing systematic near-tie splits.

\paragraph{What was invisible before.}
Prior to the diagnostic, this disagreement was attributed to annotator 
subjectivity or document ambiguity. The criterion-level analysis made 
it structurally visible: $q_5$ was simultaneously too broad and too 
ambiguous, mixing two operationally distinct signals under a single 
label. This tension was not identified during initial schema design, 
where the single-label assumption was taken as given. The diagnostic 
revealed that the problem was not in the annotators but in the 
instrument, and that the instrument itself was forcing humans to 
resolve a structural ambiguity that had never been made explicit.

\begin{figure}[ht]
\centering
\includegraphics[width=0.5\columnwidth]{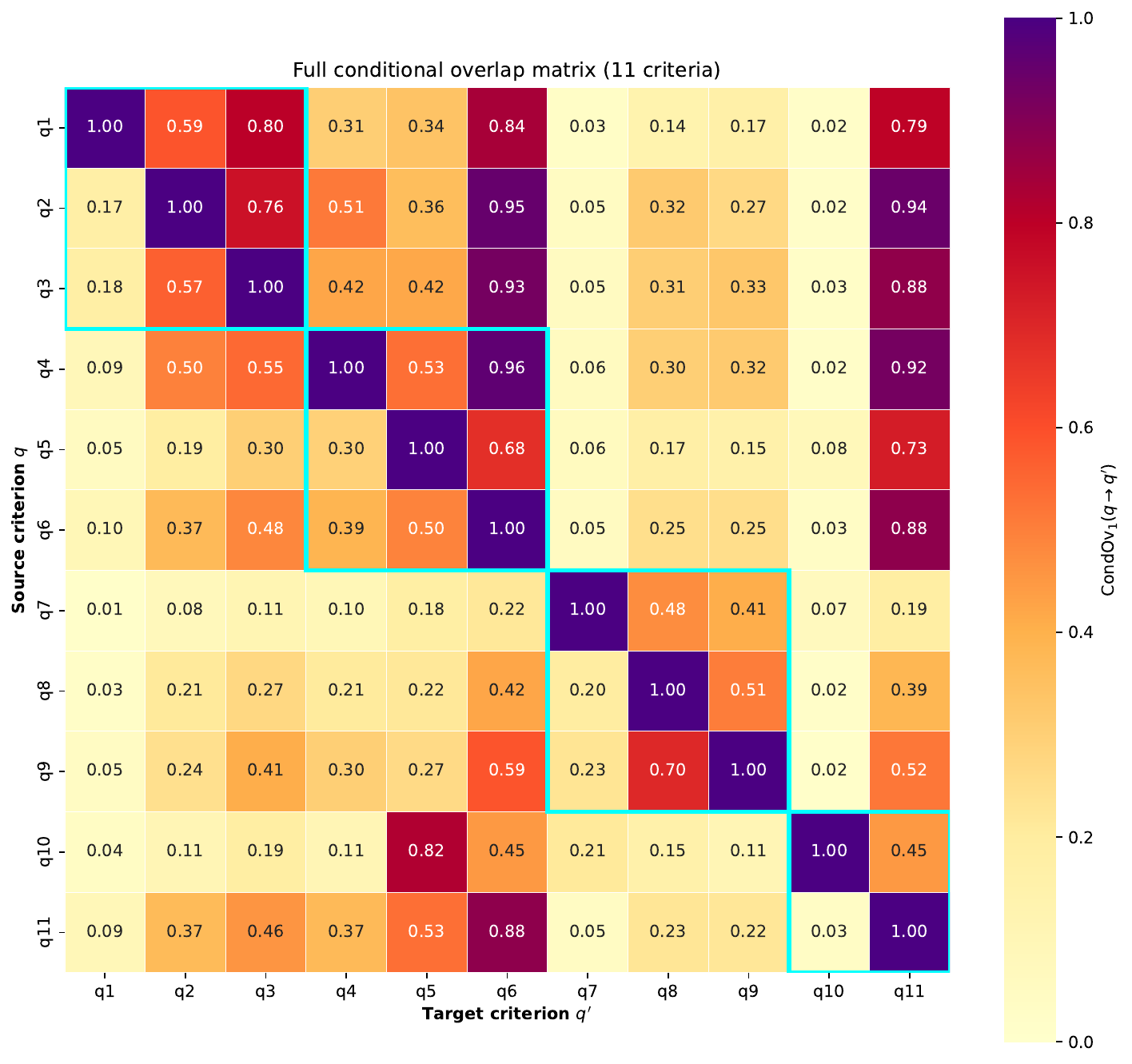}
\caption{Directed conditional overlap matrix 
$\mathrm{CondOv}_1(q\rightarrow q')$ for all 11 criteria including 
refined $q_{10}$ and $q_{11}$. Cyan borders mark within-category 
blocks. The low overlap between $q_{10}$ and $q_5$ confirms that 
the decomposition successfully isolates the explicit credibility 
signal. The moderate overlap between $q_{11}$ and $q_6$ confirms 
that $q_{11}$ extends rather than duplicates the perceived quality 
signal.}
\label{fig:overlap_refined}
\end{figure}

\paragraph{Criterion decomposition and its limits.}
Guided by these signals, domain experts decomposed $q_5$ into two 
focused criteria:

\begin{itemize}
\item $q_{10}$ (formerly $q_{5a}$): \emph{``Does the sentence explicitly 
highlight a label, certification, award, or third-party recognition?''} 
Targets verifiable credibility markers only, producing a more stable 
binary signal.

\item $q_{11}$ (merging $q_{5b}$ and $q_6$): \emph{``Does the sentence 
suggest a positive perception of the environment, usage, or user 
experience?''} Captures the subjective image and perceived quality 
signals that were previously split across two unstable criteria.
\end{itemize}
Post-refinement overlap analysis confirmed that $q_{10}$ isolates 
verifiable cases with low overlap 
($\mathrm{CondOv}(q_{10}\rightarrow q_5)=0.17$), while $q_{11}$ 
introduces a complementary rather than redundant signal 
($\mathrm{CondOv}(q_{11}\rightarrow q_6)=0.23$), as visible in 
Figure~\ref{fig:overlap_refined} which reports the full directed 
conditional overlap matrix over all 11 criteria. The detailed stability metrics for both the original and refined criteria, including activation rates and vote distributions, are provided in  Table~\ref{tab:stability_refined}

\begin{table}[ht]
\centering
\small
\setlength{\tabcolsep}{4pt}
\begin{tabular}{l l c r r r r r}
\hline
\textbf{ID} & \textbf{Criterion} & \textbf{Act$_1$ (\%)} & \textbf{NT\%} & \textbf{AS\%} & \textbf{UY\%} & \textbf{$|\Omega_{q,1}|$} \\
\hline
$q_1$  & Cost Reduction             & 2.8  & 22.9 & 45.8 & 31.3 & 131  \\
$q_2$  & Operational Efficiency     & 9.4  & 28.2 & 43.3 & 28.4 & 443  \\
$q_3$  & Organizational Impact      & 12.6 & 28.9 & 44.8 & 26.4 & 592  \\
$q_4$  & User Well-Being            & 9.8  & 35.8 & 48.3 & 15.9 & 458  \\
$q_5$  & Reputation \& Recognition  & 17.6 & 34.0 & 52.3 & 13.7 & 826  \\
$q_6$  & Tangible/Perceived Quality & 24.1 & 23.3 & 31.9 & 44.9 & 1130 \\
$q_7$  & Regulatory Compliance      & 5.8  & 25.6 & 44.7 & 29.7 & 273  \\
$q_8$  & Risk Prevention/Security   & 14.1 & 27.8 & 36.9 & 35.3 & 662  \\
$q_9$  & Mandatory Requirement      & 10.3 & 38.2 & 41.7 & 20.1 & 482  \\
\hline
$q_{10}$ & Explicit Credibility     & 1.8  & 28.6 & 54.8 & 16.7 & 84   \\
$q_{11}$ & Perceived Positive       & 24.1 & 32.0 & 33.9 & 34.1 & 1130 \\
\hline
\end{tabular}
\caption{Criterion stability at $t=1$ ($A=5$) for original ($q_1$--$q_9$) 
and refined ($q_{10}$, $q_{11}$) criteria. $q_{10}$ replaces the 
explicit credibility signal from $q_5$; $q_{11}$ merges the implicit 
image signal from $q_5$ with the perceived quality signal from $q_6$. 
Refined criteria are separated by a horizontal rule.}
\label{tab:stability_refined}
\end{table}

\begin{figure}[ht]
\centering
\includegraphics[width=0.5\columnwidth]{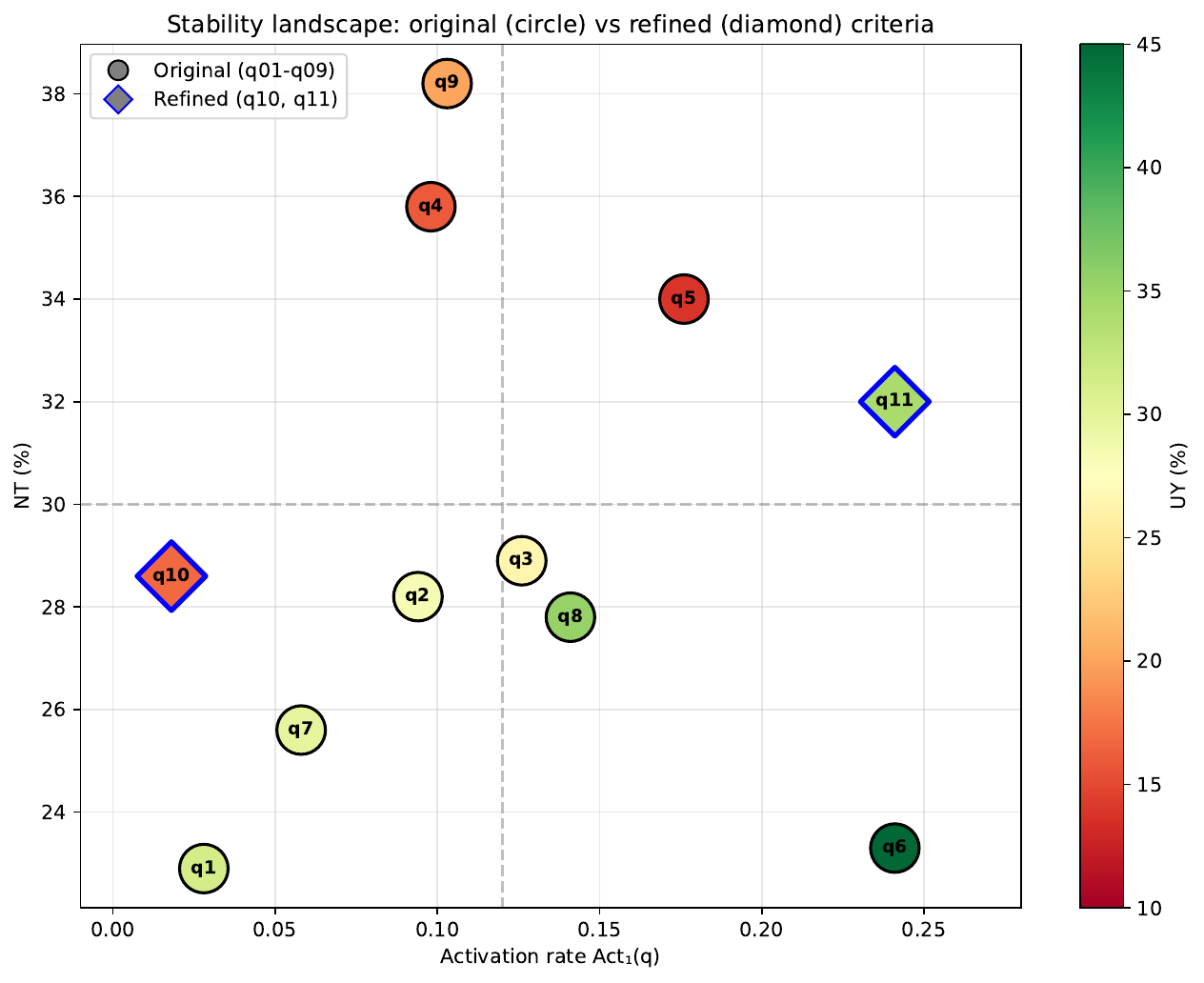}
\caption{Stability landscape for original criteria (circles, $q_1$--$q_9$) 
and refined criteria (diamonds, $q_{10}$, $q_{11}$). Color encodes 
unanimity UY. Refined criteria are outlined in blue. $q_{10}$ is rare 
but moderately stable; $q_{11}$ is frequent but more ambiguous than 
the original $q_6$ it partially replaces, reflecting the broader 
subjective scope of the merged criterion.}
\label{fig:stability_refined}
\end{figure}

However, as Figure~\ref{fig:stability_refined} illustrates, 
modifying criteria does not eliminate instability: it shifts and 
relocates it.
Changing a criterion is changing a definition, which introduces new 
semantic boundaries and new zones of ambiguity. $q_{11}$, by merging 
two previously separate signals, inherits a broader and more subjective 
scope, reflected in its position in the stability landscape. $q_{10}$, 
while more precise, activates rarely ($\mathrm{Act}_1 = 1.8\%$), 
suggesting that explicit credibility markers are sparse in this corpus. 
Refinement does not converge toward a uniquely correct schema; it 
navigates a space of design tradeoffs where each decision reshapes 
what the schema measures.

\paragraph{The structural challenge humans face.}
Beyond criterion wording, the diagnostic surfaces a deeper tension 
that cannot be resolved algorithmically. Domain experts consistently 
expressed a preference for single-label output, motivated by downstream 
procurement workflows that require mutually exclusive categories. Yet 
the corpus is inherently multi-dimensional: the same sentence routinely 
activates signals from multiple categories simultaneously. This 
mismatch between task design and content structure is not a wording 
problem but a paradigm problem.

The options under consideration reflect this tension: enforcing 
strict single-label assignment requires an explicit tie-breaking 
policy, where stakeholder-defined priorities determine which dimension 
to foreground when multiple criteria fire. This weighting would be a 
human rule, not an emergent property of the data. Alternatively, 
adopting a multi-label or hierarchical paradigm would better reflect 
the content structure but requires rethinking downstream workflows. 
Neither option is purely technical: both require normative commitments 
about what the annotation is intended to capture.

\paragraph{What the diagnostic contributes.}
The goal of schema refinement is not to maximize unanimity. High 
unanimity on a poorly specified criterion may simply mean that all 
annotators are consistently misapplying the same ambiguous rule. 
What the diagnostic provides is not a target to optimize, but a 
structured map of where disagreement concentrates and why. It 
separates specification gaps from paradigm mismatches, making 
implicit design decisions explicit and measurable, so that schema 
revision becomes an evidence-based process rather than a cycle of 
trial and error.

\section{Qualitative Examples of Diagnostic Signals}
\label{app:qual-examples}

\paragraph{Goal.}
This appendix provides qualitative examples connecting the LLM-based diagnostic to expert behavior.
We contrast (i) \emph{single-category} diagnostic cases (one induced category) with
(ii) \emph{cross-category co-activation} cases (multiple induced categories).
For each sentence, we report the diagnostic category pattern, the triggered criteria, and the distribution of expert labels
(5 experts, single-label task). Criteria identifiers $q_1$--$q_9$ follow Appendix~\ref{app:criteria}. Representative examples are provided in Table~\ref{tab:qual-examples}.

\paragraph{Category key.}
C0 = Description (non-persuasive),\,
C1 = Performance \& Efficiency,\,
C2 = User Experience \& Brand Value,\,
C3 = Obligation \& Safety.

\begin{table}[th]
\centering
\small
\setlength{\tabcolsep}{4pt}
\renewcommand{\arraystretch}{1.15}
\resizebox{\textwidth}{!}{
\begin{tabular}{p{0.9cm} p{9.2cm} p{2.3cm} p{2.3cm} p{3.4cm} p{2.6cm}}
\hline
\textbf{ID} &
\textbf{Sentence (EN translation)} &
\textbf{Diagnostic pattern} &
\textbf{Criteria (LLM)} &
\textbf{Expert labels ($n=5$)} &
\textbf{Signal}\\
\hline

\textbf{708} &
Our agreement specifies GDPR-compliant data processing between you, us, and the employer providing the data. &
$c_3$ &
$q_7, q_8, q_9$ &
$c_3\!:\!5$ &
Single-category (easy) \\

\textbf{1132} &
To minimize risks related to the customer's application, adequate safeguards must be implemented to reduce hazards. &
$c_3$ &
$q_8, q_9$ &
$c_3\!:\!4,\; c_0\!:\!1$ &
Mostly single-category \\

\textbf{1011} &
Designed to generate savings and be profitable. &
$c_1{+}c_2$ &
$q_1, q_2, q_3, q_6$ &
$c_1\!:\!4,\; c_0\!:\!1$ &
$c_1$--$c_2$ split \\

\textbf{2238} &
Its role is to improve farm performance while promoting sustainable development of local agriculture. &
$c_1{+}c_2$ &
$q_2, q_3, q_6$ &
$c_1\!:\!3,\; c_2\!:\!2$ &
$c_1$--$c_2$ split \\

\textbf{1345} &
Monitor outdoor air quality: alert when levels exceed standards. &
$c_2{+}c_3$ &
$q_6, q_7, q_8, q_9$ &
$c_2\!:\!1,\; c_1\!:\!2,\; c_3\!:\!2$ &
Cross-boundary (mixed) \\

\textbf{199} &
They integrate innovative technologies to improve the security posture. &
$c_1{+}c_2{+}c_3$ &
$q_2, q_3, q_6, q_8, q_9$ &
$c_3\!:\!4,\; c_2\!:\!1$ &
Multi-category (hard) \\

\hline
\end{tabular}
}
\caption{Qualitative examples linking diagnostic flags to expert splits.
``Diagnostic pattern'' reports induced category activity, and ``Criteria (LLM)'' lists triggered criterion IDs $q_1$--$q_9$ (Appendix~\ref{app:criteria}).
Expert labels are summarized as counts over five experts (single-label task).
Cross-category cases align with expert disagreement, whereas single-category $c_3$ cases are largely stable.}
\label{tab:qual-examples}
\end{table}


\end{document}